\documentclass[10pt]{sigplanconf}

\usepackage{amsmath}

\usepackage[normalem]{ulem}
\usepackage{color,soul}
\usepackage{multirow}
\usepackage{amsmath}
\usepackage[ruled,vlined,linesnumbered]{algorithm2e}
\usepackage{graphicx} 
\renewcommand\footnoterule{}

\usepackage{authblk}
\usepackage[]{natbib}
\usepackage{url}
\setcitestyle{numbers}

\begin{document}
\CopyrightYear{2017} 
\setcopyright{acmcopyright}
\conferenceinfo{ASPLOS '17,}{April 08-12, 2017, Xi'an, China}
\isbn{978-1-4503-4465-4/17/04}\acmPrice{\$15.00}
\doi{http://dx.doi.org/10.1145/3037697.3037746}

\title{SC-DCNN: Highly-Scalable Deep Convolutional \\Neural Network using Stochastic Computing}

\authorinfo{ Ao Ren$^{\dagger}$ ~ Ji Li$^\diamond$ ~ Zhe Li$^\dagger$ ~Caiwen Ding$^\dagger$ \\ Xuehai Qian$^\diamond$ ~ Qinru Qiu$^\dagger$ ~ Bo Yuan$^\mp$ ~ Yanzhi Wang$^{\dagger}$ \\ {\it $^\dagger$ Department of Electrical Engineering and Computer Science, Syracuse University \\~ $^\diamond$Department of Electrical Engineering, University of Southern California \\~ $^\mp$Department of Electrical Engineering, City University of New York, City College} \\ {\it $^\dagger$ \{aren, zli89, cading, qiqiu, ywang393\}@syr.edu, \\ ~ $^\diamond$ \{jli724, xuehai.qian\}@usc.edu, ~ $^\mp$ byuan@ccny.cuny.edu}\vspace{-3ex}}
	

\setlength{\belowcaptionskip}{-10pt}

\setlength{\emergencystretch}{3em}
\date{}
\maketitle

\begin{abstract}
\vskip -0.1em
With the recent advance of wearable devices and Internet of Things (IoTs), it becomes attractive to implement the Deep Convolutional Neural Networks (DCNNs) in embedded and portable systems. 
Currently, executing the software-based DCNNs requires high-performance servers, 
restricting the widespread deployment on embedded and mobile IoT devices. 
To overcome this obstacle, considerable research efforts have 
been made to develop highly-parallel and specialized DCNN accelerators using GPGPUs, FPGAs or ASICs.



Stochastic Computing (SC), which uses a bit-stream to represent a number within [-1, 1] by counting the number of ones in the bit-stream, has high potential for implementing DCNNs with high scalability and ultra-low hardware footprint. 
Since multiplications and additions can be calculated using AND gates and multiplexers in SC, significant reductions in power (energy) and hardware footprint can be achieved compared to the conventional binary arithmetic implementations. 
The tremendous savings in power (energy) and hardware resources allow immense design space for enhancing scalability and robustness for hardware DCNNs.

This paper presents SC-DCNN, the first comprehensive design and optimization framework of SC-based DCNNs, 
using a bottom-up approach.
We first present the designs of function blocks that perform the basic operations in DCNN, including
inner product, pooling, and activation function. 
Then we propose four designs of feature extraction blocks, which are in charge of 
extracting features from input feature maps, by connecting different basic function blocks
with joint optimization. 
Moreover, the efficient weight storage methods 
are proposed to reduce the area and power (energy) consumption. 
Putting all together, with feature extraction blocks carefully selected, 
SC-DCNN is holistically optimized to minimize area and power (energy) consumption 
while maintaining high network accuracy. Experimental results demonstrate 
that the LeNet5 implemented in SC-DCNN consumes only 17 $mm^2$ area and 1.53 W power, 
achieves throughput of 781250 images/s, area efficiency of 45946 images/s/$mm^2$, and energy efficiency of 510734 images/J.

\end{abstract}

\section{Introduction}
\vskip -0.25em
Deep learning (or deep structured learning) has emerged as a new area of machine learning research, which enables a system to automatically learn complex information and extract representations at multiple levels of abstraction \cite{deng2014deep}.
\textit{Deep Convolutional Neural Network (DCNN)} is recognized as 
one of the most promising types of artificial neural networks taking advantage of deep learning and
has become the dominant approach for almost all recognition and detection tasks \cite{lecun2015deep}.
Specifically, DCNN has achieved significant success in a wide range of machine learning applications, such as image classification \cite{simonyan2014very}, natural language processing \cite{collobert2008unified}, speech recognition \cite{sainath2013deep}, and video classification \cite{karpathy2014large}.

Currently, the high-performance servers are usually required for executing software-based DCNNs since software-based DCNN implementations involve a large amount of computations to achieve outstanding performance.
However, the high-performance servers incur high power (energy) consumption and large hardware cost, 
making them unsuitable for applications in embedded and mobile IoT devices that 
require low-power consumption.
These applications play an increasingly important role in our everyday life and exhibit a notable trend of being ``smart''.
To enable DCNNs in these application with low-power and low-hardware cost, 
the highly-parallel and specialized hardware has been designed using 
General-Purpose Graphics Processing Units (GPGPUs), Field-Programmable Gate Array (FPGAs), 
and Application-Specific Integrated Circuit (ASICs)~\cite{laszlo2012analysis, stoica2015high, zhang2015optimizing, motamedi2016design, andri2016yodann, tanomoto2015cgra, akopyan2015truenorth, strukov2008missing, hu2014memristor, xia2016mnsim, chen2014dadiannao, EIE}.
Despite the performance and power (energy) efficiency achieved, 
a large margin of improvement still exists due to the inherent inefficiency in implementing DCNNs using conventional computing methods or using general-purpose computing devices \cite{ji2015hardware,kim2016dynamic}.  


We consider \emph{Stochastic Computing (SC)} as a novel computing paradigm
to provide significantly low hardware footprint with high energy efficiency and scalability. 
In SC, a probability number is represented using a bit-stream \cite{gaines1967stochastic}, therefore,
the key arithmetic operations such as multiplications and additions can be implemented as simple as AND gates and multiplexers (MUX), respectively \cite{brown2001stochastic}.
Due to these features, SC has 
the potential to implement DCNNs with significantly reduced hardware resources and high power (energy) efficiency. 
Considering the large number of multiplications and additions in DCNN, 
achieving the efficient DCNN implementation using SC requires the exploration 
of a large design space.

In this paper, we propose {\em SC-DCNN}, 
the first comprehensive design and optimization framework of SC-based DCNNs, using a bottom-up approach.

The proposed SC-DCNN fully utilizes the advantages of SC and achieves remarkably low hardware footprint, 
low power and energy consumption, while maintaining high network accuracy. 
Based on the proposed SC-DCNN architecture, this paper made the following key contributions:

\begin{itemize}
\item \textbf{Applying SC to DCNNs.} 
We are the {\em first} (to the best of our knowledge) to apply SC to DCNNs.
This approach is motivated by {\em 1)} the potential of SC as a computing paradigm to 
provide low hardware footprint with high energy efficiency and scalability;
and {\em 2)} the need to implement DCNNs in the embedded and mobile IoT devices. 

\item \textbf{Basic function blocks and hardware-oriented max pooling.} 
We propose the design of \emph{function blocks} that perform the basic operations in DCNN, including
Specifically, we present a novel hardware-oriented max pooling design for 
efficiently implementing (approximate) max pooling in SC domain. 
The pros and cons of different types of function blocks are also thoroughly investigated.

\item \textbf{Joint optimizations for feature extraction blocks.} 
We propose four optimized designs of {\em feature extraction blocks} 
which are in charge of extracting features from input feature maps. 
The function blocks inside the feature extraction block are {\em jointly optimized} 
through both analysis and experiments with respect to input bit-stream length, 
function block structure, and function block compatibilities.

\item \textbf{Weight storage schemes.} 
The area and power (energy) consumption of weight storage are
reduced by comprehensive techniques, including 
efficient filter-aware SRAM sharing, effective weight storage methods, and layer-wise weight storage optimizations.

\item \textbf{Overall SC-DCNN optimization.} 
We conduct holistic optimizations of the overall SC-DCNN architecture with carefully selected feature extraction blocks and layer-wise feature extraction block configurations,
to minimize area and power (energy) consumption while maintaining the high
network accuracy. 
The optimization procedure leverages the crucial observation that hardware inaccuracies in different layers 
in DCNN have different effects on the overall accuracy.
Therefore, different designs may be exploited to minimize area and power (energy) consumptions.

\item \textbf{Remarkably low hardware footprint and low power (energy) consumption.} 
Overall, the proposed SC-DCNN achieves the {\em lowest} hardware cost and energy consumption in implementing LeNet5 compared with reference works.
\end{itemize}

\section{Related Works}
\vskip -0.25em
Authors in \cite{laszlo2012analysis, stoica2015high, krizhevsky2012imagenet, jia2014caffe} leverage the parallel computing and storage resources in GPUs for efficient DCNN implementations.
FPGA-based acceleration~\cite{zhang2015optimizing, motamedi2016design} is another promising path towards the hardware implementation of DCNNs
due to the programmable logic, high degree of parallelism and short develop round. 
However, the GPU and FPGA-based implementations still exhibit a large margin of performance enhancement and power reduction. 
It is because {\em 1)} GPUs and FPGAs are general-purpose computing devices not specifically optimized for executing DCNNs, and {\em ii)} the relatively limited signal routing resources in such general platforms restricts 
the performance of DCNNs which typically exhibit high inter-neuron communication requirements.

ASIC-based implementations of DCNNs have been recently exploited to overcome the limitations of general-purpose computing devices. Two representative recent works are DaDianNao \cite{chen2014dadiannao} and EIE \cite{EIE}. 
The former proposes an ASIC ``node'' which could be connected in parallel to implement a large-scale DCNN, whereas the latter focuses specifically on the fully-connected layers of DCNN and achieves high throughput and energy efficiency.  

To significantly reduce hardware cost and improve energy efficiency and scalability, 
novel computing paradigms need to be investigated.
We consider SC-based implementation of neural network an attractive candidate 
to meet the stringent requirements and facilitate the 
widespread of DCNNs in embedded and mobile IoT devices. 
Although not focusing on deep learning, \cite{sato2003implementation} proposes the design of a neurochip using stochastic logic.
\cite{ji2015hardware} utilizes stochastic logic to implement a radial basis function-based neural network.
In addition, a neuron design with SC for deep belief network was presented in \cite{kim2016dynamic}.
Despite the previous application of SC, there is no existing work that investigates comprehensive designs and optimizations of SC-based hardware DCNNs including both computation blocks and weight storing methods. 

\section{Overview of DCNN Architecture and Stochastic Computing}
\vskip -0.25em
\subsection{\textit{DCNN Architecture Overview}}

\begin{figure}[b]
	\centering
	\includegraphics[width=0.9\columnwidth]{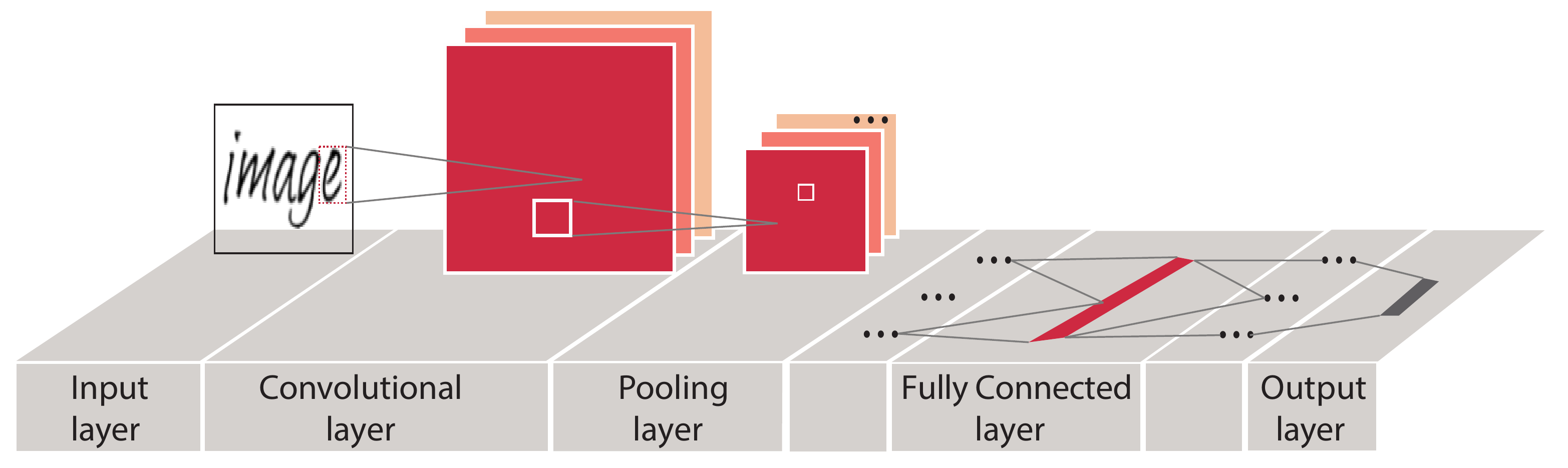}
	\caption{The general DCNN architecture.}
	\label{fig:dcnn_overview}
\end{figure}

Deep convolutional neural networks are biologically inspired variants of multilayer perceptrons (MLPs) by mimicking the animal visual mechanism \cite{lenet}. An animal visual cortex contains two types of cells and they are only sensitive to a small region (receptive field) of the visual field. Thus a neuron in a DCNN is only connected to a small receptive field of its previous layer, rather than connected to all neurons of previous layer like traditional fully connected neural networks.

\begin{figure}[t]
	\centering
	\includegraphics[width=0.8\columnwidth]{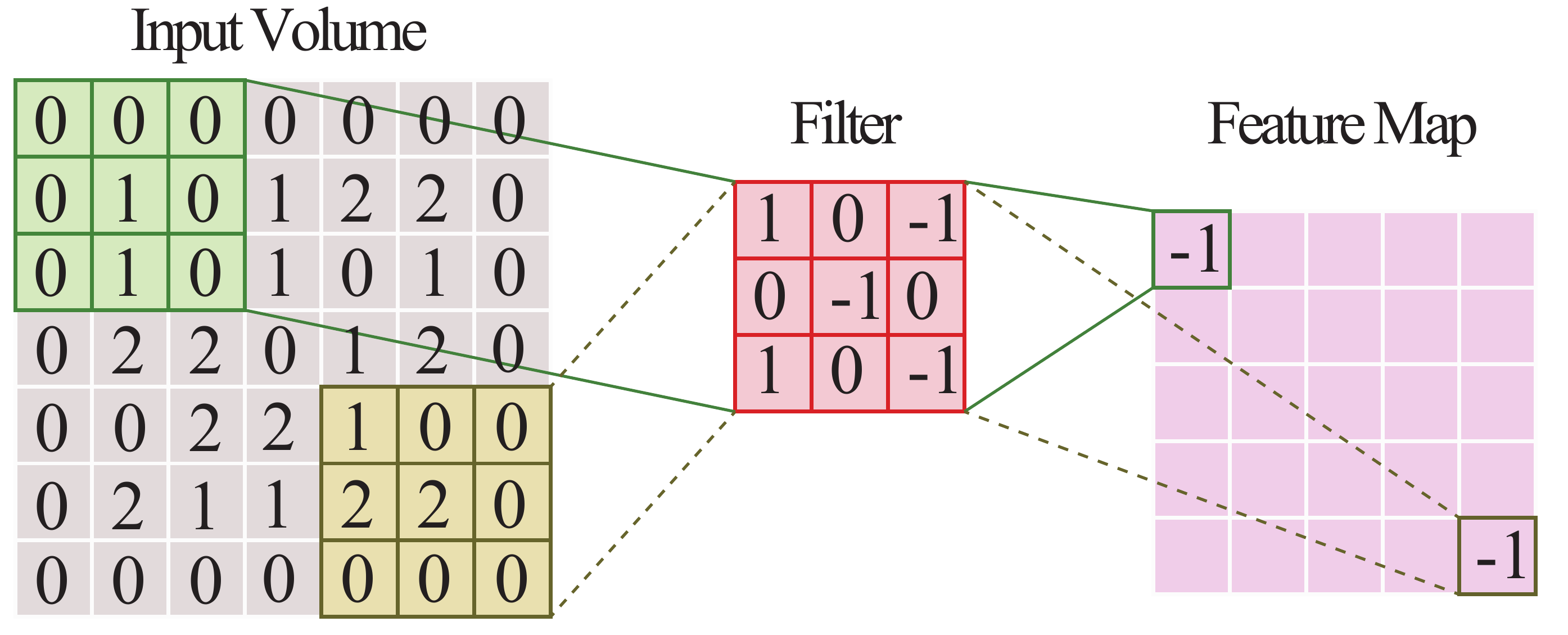}
	\caption{Illustration of the convolution process.}
	\label{fig:convolution}
\end{figure}

As shown in Figure \ref{fig:dcnn_overview}, in the simplest case, 
a DCNN is a stack of three types of layers: \emph{Convolutional Layer}, \emph{Pooling Layer}, and \emph{Fully Connected Layer}. The Convolutional layer is the core building block of DCNN, its main operation is the convolution that calculates the dot-product of receptive fields and a set of learnable filters (or kernels) \cite{cs231n}. Figure \ref{fig:convolution} illustrates the process of convolution operations. After the convolution operations, the nonlinear down-samplings are conducted in the pooling layers to reduce the dimension of data. The most common pooling strategies are \emph{max pooling} and \emph{average pooling}. Max pooling picks up the maximum value from the candidates, and average pooling calculates the average value of the candidates. Then, the extracted feature maps after down-sampling operations are sent to \emph{activation functions} that conduct non-linear transformations such as Rectified Linear Unit (ReLU) $ f(x)=max(0,x)$, Sigmoid function $ f(x)=(1+e^{-x}){-1} $ and hyperbolic tangent (tanh) function $ f(x)= \frac{2}{1+e^{-2x}}-1$. Next, the high-level reasoning is completed via the fully connected layer. Neurons in this layer are connected to all activation results in the previous layer. Finally, the loss layer is normally the last layer of DCNN and it specifies how the deviation between the predicted and true labels is penalized in the network training process. Various loss functions such as softmax loss, sigmoid cross-entropy loss may be used for different tasks. 

\begin{figure}[t]
	\centering
	\includegraphics[width=0.95\columnwidth]{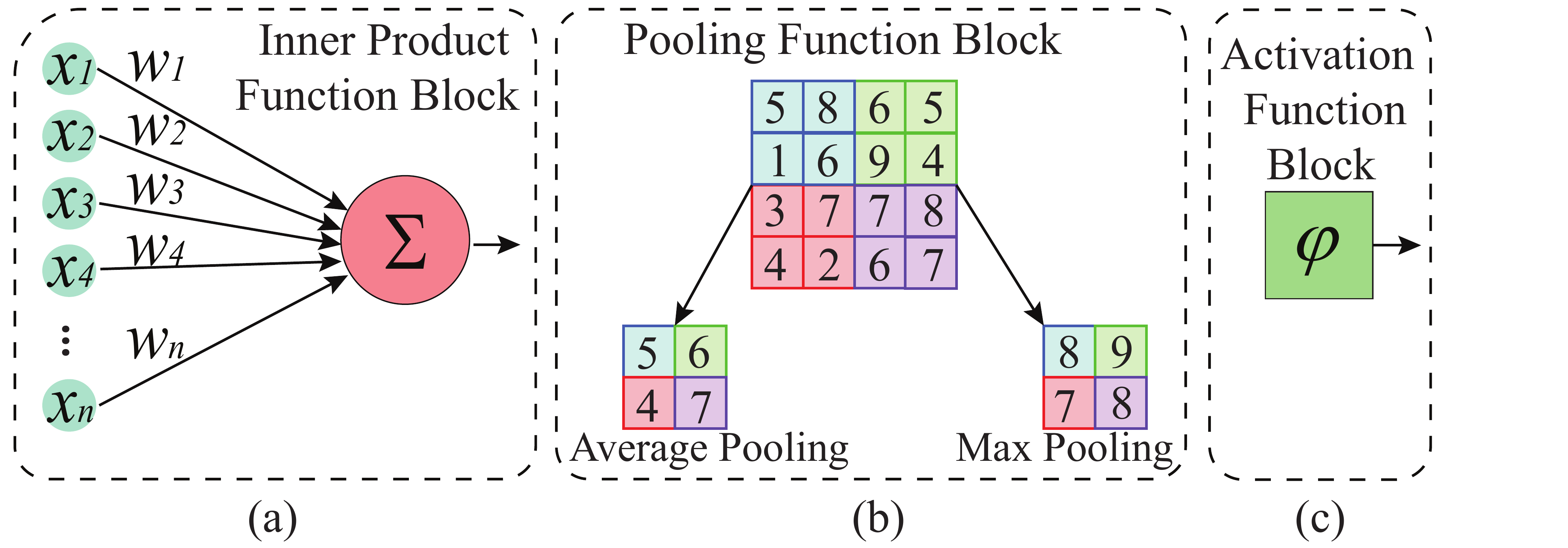}
	\caption{Three types of basic operations (function blocks) in DCNN. (a) Inner Product, (b) pooling, and (c) activation. }
	\label{fig:neuronconcept}
\end{figure}

The concept of ``neuron'' is widely used in the software/algorithm domain. In the context of DCNNs, a neuron consists
of one or multiple basic operations. 
In this paper, we focus on the basic operations in hardware designs and optimizations, including: 
inner product, pooling, and activation. The corresponding SC-based designs of these fundamental operations are termed \emph{function blocks}. Figure \ref{fig:neuronconcept} illustrates the behaviors of function blocks, where $x_i$'s in Figure \ref{fig:neuronconcept}(a) represent the elements in a receptive filed, and $w_i$'s represent the elements in a filter.  
Figure \ref{fig:neuronconcept}(b) shows the average pooling and max pooling function blocks. 
Figure \ref{fig:neuronconcept}(c) shows the activation function block (e.g. hyperbolic tangent function). 
The composition of an inner product block, a pooling block, and an activation function block is referred to as the \emph{feature extraction block}, which extracts features from feature maps. 


\subsection{Stochastic Computing (SC)}
\label{sc_basic}
Stochastic Computing (SC) is a paradigm that represents a probabilistic number by counting the number of ones in a bit-stream. For instance, the bit-stream 0100110100 contains four ones in a ten-bit stream, thus it represents $ P(X=1)=4/10=0.4 $. In addition to this unipolar encoding format, SC can also represent numbers in the range of [-1, 1] using the bipolar encoding format. In this scenario, a real number $ x $ is processed by $ P(X=1)=(x+1)/2 $, thus 0.4 can be represented by 1011011101, as $ P(X=1)=(0.4+1)/2=7/10 $. To represent a number beyond the range [0, 1] using unipolar format or beyond [-1, 1] using bipolar format, a pre-scaling operation \cite{yuan2016design} can be used. Furthermore, since the bit-streams are randomly generated with stochastic number generators (SNGs),
the randomness and length of the bit-streams can significantly affect the calculation accuracy~\cite{aozhe2016icrc}. 
Therefore, the efficient utilization of SNGs and the trade-off between the bit-stream length (i.e. the accuracy) and the resource consumption need to be carefully taken into consideration.

Compared to the conventional binary computing, 
the major advantage of stochastic computing is the significantly lower hardware cost 
for a large category of arithmetic calculations.
The abundant area budget offers immense design space in optimizing hardware performance via 
exploring the tradeoffs between the area and other metrics, such as power, latency, and parallelism degree.
Therefore, SC is an interesting and promising approach to implementing large-scale DCNNs.

\begin{figure}[t]
	\centering
	\includegraphics[width=0.78\columnwidth]{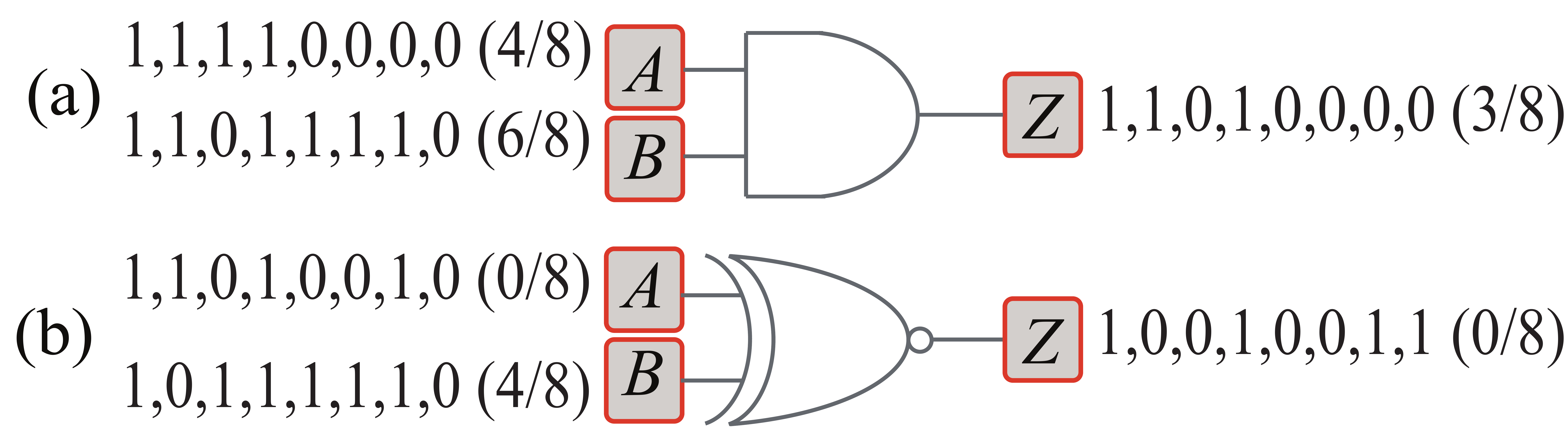}
	\caption{Stochastic multiplication. (a) Unipolar multiplication and (b) bipolar multiplication.}
	\label{fig:multiplication}
\end{figure}

\textbf{Multiplication}. Figure \ref{fig:multiplication} shows the basic multiplication components in SC domain. A unipolar multiplication can be performed by an AND gate since $ P(A \cdot B=1)=P(A=1)P(B=1) $ (assuming independence of two random variables), and a bipolar multiplication is performed by means of a XNOR gate since $ c=2P(C=1)-1=2(P(A=1)P(B=1)+P(A=0)P(B=0))-1=(2P(A=1)-1)(2P(B=1)-1)=ab $. 

\begin{figure}[t]
	\centering
	\includegraphics[width=0.8\columnwidth]{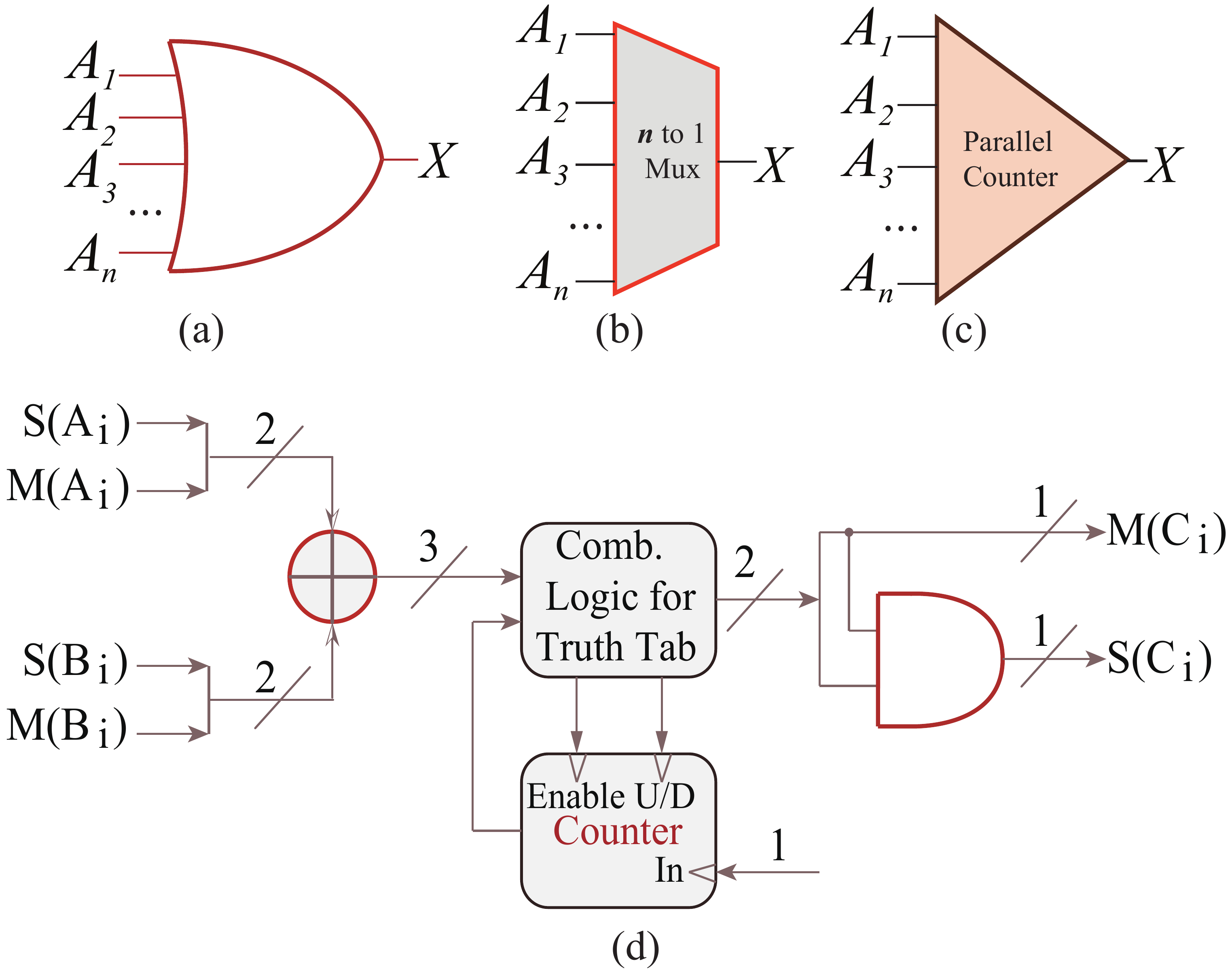}
	\caption{Stochastic addition. (a) OR gate, (b) MUX, (c) APC, and (d) two-line representation-based adder.}
	\label{fig:sc_addition}
\end{figure}

\textbf{Addition}. 
We consider four popular stochastic addition methods for SC-DCNNs. 
{\em 1)} OR gate (Figure~\ref{fig:sc_addition} (a)). It is the simplest method that consumes the least hardware footprint to perform an addition, but this method introduces considerable accuracy loss because the computation ``logic 1 OR logic 1" only generates a single logic 1. 
{\em 2)} Multiplexer (Figure \ref{fig:sc_addition} (b)). It uses a multiplexer, which is the most popular way to perform additions in either the unipolar or the bipolar format \cite{brown2001stochastic}. 
For example, a bipolar addition is performed as 
$c=2P(C=1)-1=2(1/2P(A=1)+1/2P(B=1))-1=1/2(2P(A=1)-1)+(2P(B=1)-1))=1/2(a+b)$. 
{\em 3)} Approximate parallel counter (APC)~\cite{kim2015approximate}  
(Figure \ref{fig:sc_addition} (c)). It counts the number of 1s in the inputs and represents the result with a binary number. This method consumes fewer logic gates compared with the conventional accumulative parallel counter \cite{kim2015approximate, parhami1995accumulative}. 
{\em 4)} Two-line representation of a stochastic number~\cite{toral2000stochastic} 
(Figure \ref{fig:sc_addition} (d)). This representation consists of a magnitude stream $ M(X) $ and a sign stream $ S(X) $, in which 1 represents a negative bit and 0 represents a positive bit. The value of the represented stochastic number is calculated by: $ x=\frac{1}{L} \sum_{i=0}^{L-1} (1-2S(X_{i}))M(X_{i}) $, where $ L $ is the length of the bit-stream. As an example, -0.5 can be represented by $ M(-0.5): 10110001 $ and $ S(-0.5): 11111111 $. 
Figure \ref{fig:sc_addition} (d) illustrates the structure of the two-line representation-based adder. The summation of $A_i$ (consisting of $S(A_i)$ and $M(A_i)$) and $B_i$ are sent to a truth table, then the truth table and the counter together determine the carry bit and $C_i$. The truth table can be found in \cite{toral2000stochastic}.  

\begin{figure}[t]
	\centering
	\includegraphics[width=0.8\columnwidth]{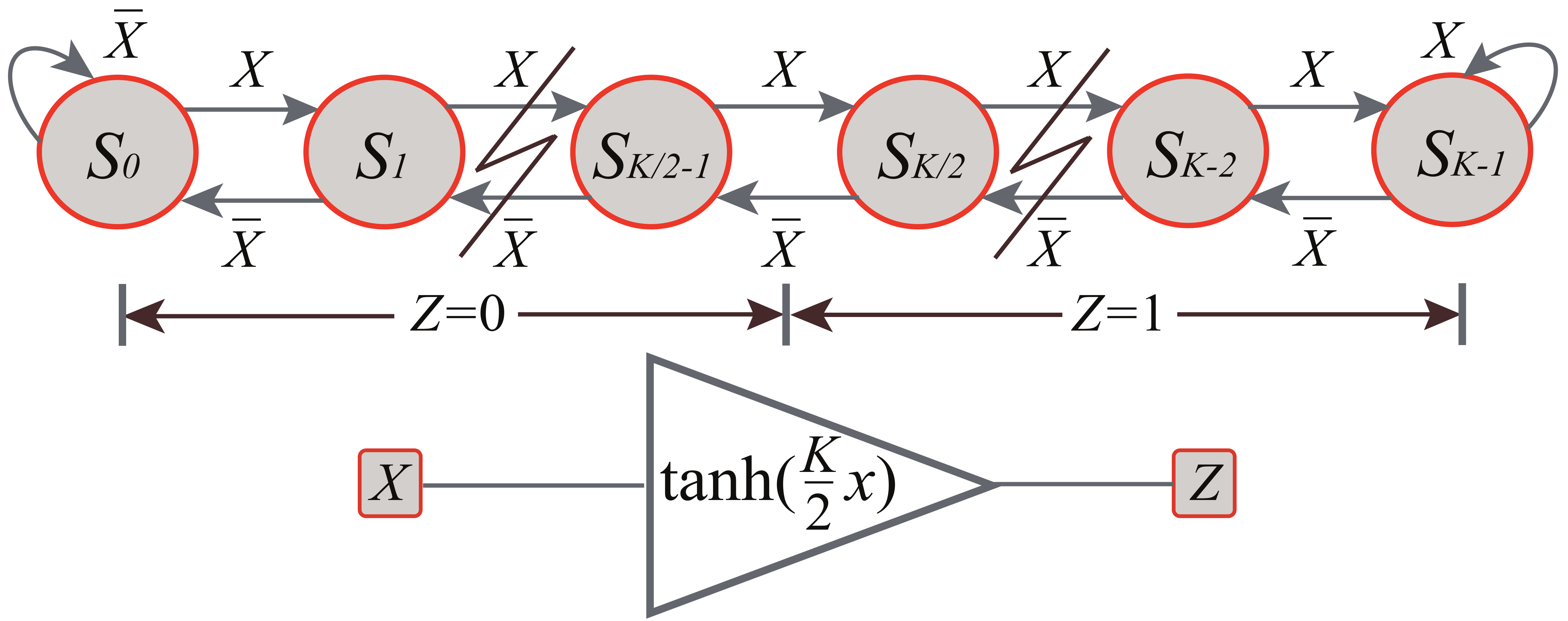}
	\caption{Stochastic hyperbolic tangent.}
	\label{fig:tanh_fsm}
\end{figure}

\textbf{Hyperbolic Tangent (tanh)}. The tanh function is highly suitable for SC-based implementations because {\em i)} it can be easily implemented with a K-state finite state machine (FSM) in the SC domain \cite{brown2001stochastic, li2017towards} and costs less hardware when compared to the piecewise linear approximation (PLAN)-based implementation \cite{larkin2006efficient} in conventional computing domain; and {\em ii)} replacing ReLU or sigmoid function by tanh function does not cause accuracy loss in DCNN \cite{krizhevsky2012imagenet}. Therefore, we choose tanh as the activation function in SC-DCNNs in our design. 
The diagram of the FSM is shown in Figure \ref{fig:tanh_fsm}. It reads the input bit-stream bit by bit, when the current input bit is one, it moves to the next state, otherwise it moves to the previous state. It outputs a 0 when the current state is on the left half of the diagram, otherwise it outputs a 1. The value calculated by the FSM satisfies $ Stanh(K, x) \cong tanh(\frac{K}{2}x) $, where $Stanh$ denotes stochastic tanh.

\subsection{Application-level vs. Hardware Accuracy}
The overall network accuracy 
(e.g., the overall recognition or classification rates) is one of the key optimization goals of the SC-based hardware DCNN. 
Due to the inherent stochastic nature,
the SC-based function blocks and feature extraction blocks exhibit certain degree of hardware inaccuracy. 
The network accuracy and hardware accuracy are different but correlated, --- the high accuracy in each function block will likely lead to a high overall 
network accuracy. 
Therefore, the hardware accuracy can be optimized in the design of SC-based function blocks and feature extraction blocks.

\section{Design and Optimization for Function Blocks and Feature Extraction Blocks in SC-DCNN} \label{sctn3_neurondesign}

In this section, we first perform comprehensive designs and optimizations in order to derive the most efficient SC-based implementations for function blocks, including inner product/convolution, pooling, and activation function.
The goal is to reduce power, energy and hardware resource while 
still maintaining high accuracy. Based on the detailed analysis of pros and cons of each basic function block design, we propose the optimized designs of feature extraction blocks for SC-DCNNs through both analysis and experiments. 


\subsection{Inner Product/Convolution Block Design}
As shown in Figure \ref{fig:neuronconcept} (a), an inner product/convolution block in DCNNs is composed of multiplication and addition operations. In SC-DCNNs, 
inputs are distributed in the range of [-1, 1], we adopt the bipolar multiplication implementation (i.e. XNOR gate) for the inner product block design. The summation of all products is performed by the adder(s). 
As discussed in Section~\ref{sc_basic}, the addition operation has different implementations.
To find the best option for DCNN, we replace the 
summation unit in Figure \ref{fig:neuronconcept} (a) with the four different adder implementations shown in Figure \ref{fig:sc_addition}.

\textbf{OR Gate-Based Inner Product Block Design}. 
Performing addition using OR gate is straightforward. 
For example, $\frac{3}{8} + \frac{4}{8}$ can be performed by 
"00100101 OR 11001010", which generates "11101111" $(\frac{7}{8})$. However, if first input bit-stream is changed to "10011000", the output of OR gate becomes "11011010" $(\frac{5}{8})$.
Such inaccuracy is introduced by the multiple representations of the 
same value in SC domain and the fact that the simple "logic 1 OR logic 1" 
cannot tolerate 
such variance. To reduce the accuracy loss, the input streams should be pre-scaled to ensure that there are only very few 1's in the bit-streams. For the unipolar format bit-streams, the scaling can be easily done by dividing the original number by a scaling factor. Nevertheless, in the scenario of bipolar encoding format, there are about 50\% 1's in the bit-stream when the original value is close to 0. 
It renders the scaling ineffective in reducing the number of 1's in the bit-stream.  

Table \ref{tbl_orgate} shows the average inaccuracy (absolute error) of OR gate-based inner product block with different input sizes, in which the bit-stream length is fixed at 1024 and all average inaccuracy values are obtained with the most suitable pre-scaling. The experimental results suggest that the accuracy of unipolar calculations may be acceptable, but the accuracy is too low for bipolar calculations and becomes even worse with the increased input size. Since it is almost impossible to have only positive input values and weights, 
we conclude that the OR gate-based inner product block is not appropriate for SC-DCNNs.

\begin{table}[tbp]\footnotesize
\centering
\caption{Absolute Errors of OR Gate-Based Inner Product Block}
\label{tbl_orgate}
\begin{tabular}{c c c c c}
\hline
Input Size      & 16   & 32   & 64   \\ \hline
Unipolar inputs & 0.47 & 0.66 & 1.29 \\ 
Bipolar inputs  & 1.54 & 1.70 & 2.3 \\ \hline
\end{tabular}
\end{table}

\textbf{MUX-Based Inner Product Block Design}. According to \cite{brown2001stochastic}, an $n$-to-1 MUX can sum all inputs together and generate an output with a scaling down factor $\frac{1}{n}$. Since only one bit is selected among all inputs to that MUX at one time, the probability of each input to be selected is $\frac{1}{n}$. The selection signal is controlled by a randomly generated natural number between 1 and $n$. Taking Figure \ref{fig:neuronconcept} (a) as an example, the output of the summation unit (MUX) is $ \frac{1}{n}(x_{0}w_{0}+...+x_{n-1}w_{n-1}) $.

Table \ref{tbl_muxinner} shows the average inaccuracy (absolute error) of the MUX-based inner product block measured with different input sizes and bit-stream lengths. The accuracy loss of MUX-based block is mainly caused by the fact that only one input is selected at one time, and all the other inputs are not used. The increasing input size causes accuracy reduction because more bits are dropped.
However, we see that sufficiently good accuracy can still be obtained by increasing the bit-stream length.  

\begin{table}[tbp]\footnotesize
\centering
\caption{Absolute Errors of MUX-Based Inner Product Block}
\label{tbl_muxinner}
\begin{tabular}{ccccc}
\hline
 \multirow{2}{*}{Input size} & 	 \multicolumn{4}{c}{\begin{tabular}[c]{@{}c@{}}Bit stream length\end{tabular}} \\ \cline{2-5}
  & 512  & 1024 & 2048 & 4096 \\ \hline
16                              & 0.54 & 0.39 & 0.28 & 0.21 \\ 
32                              & 1.18 & 0.77 & 0.56 & 0.38 \\
64                              & 2.35 & 1.58 & 1.19 & 0.79 \\ \hline
\end{tabular}
\end{table}

\begin{figure}[t]
	\centering
	\includegraphics[width=0.8\columnwidth]{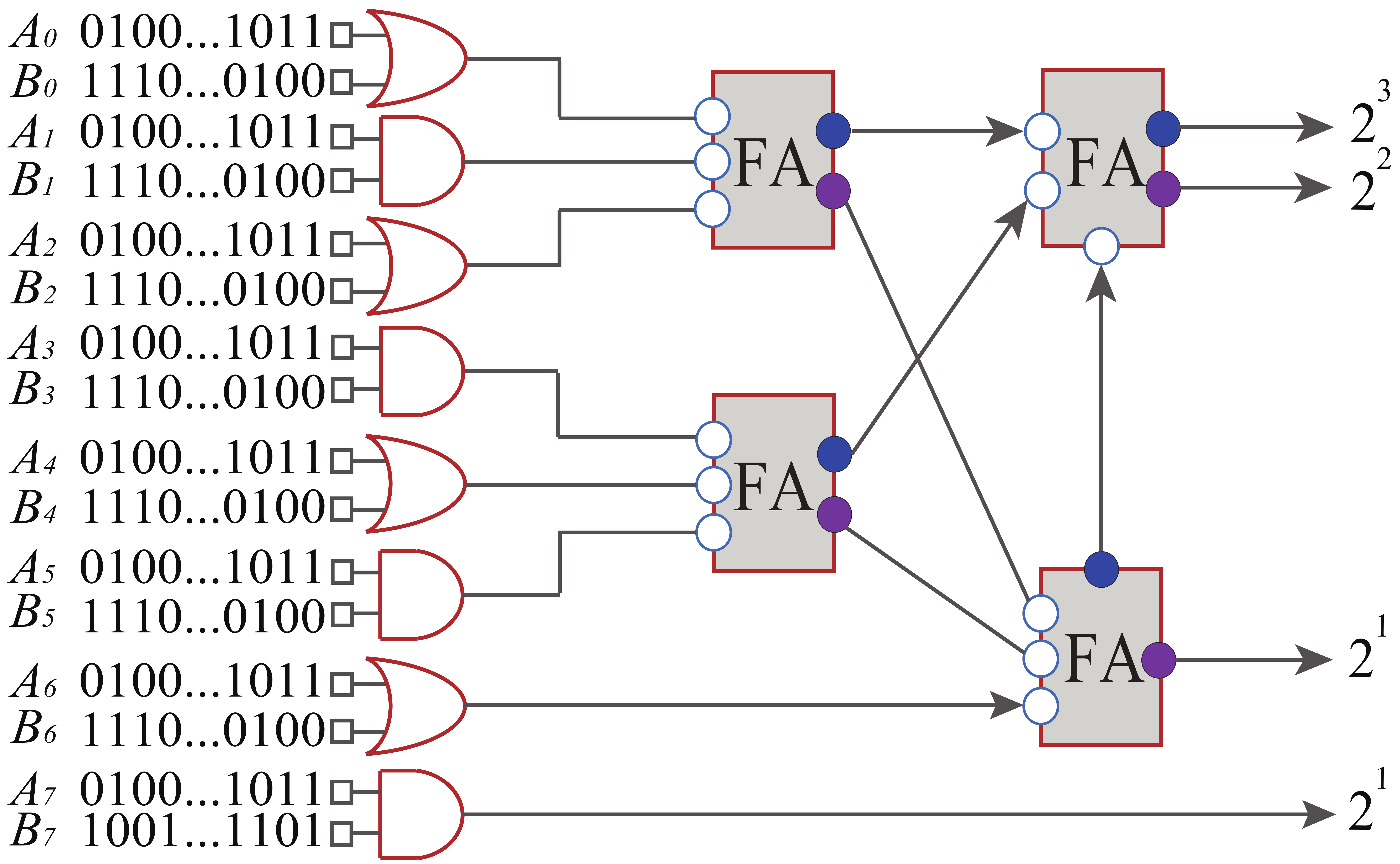}
	\caption{16-bit Approximate Parallel Counter.}
	\label{fig:16bit_apc}
\end{figure}

\textbf{APC-Based Inner Product Block}. The structure of a 16-bit APC is shown in Figure \ref{fig:16bit_apc}. $A_{0}-A_{7}$ and $B_{0}-B_{7}$ are the outputs of XNOR gates, i.e., the products of inputs $x_{i}$'s and weights $w_{i}$'s. Suppose the number of inputs is $n$ and the length of a bit-stream is $m$, then the products of $x_{i}$'s and $w_{i}$'s can be represented by a bit-matrix of size $n\times m$. The function of the APC is to count the number of ones in each column and represent the result in the binary format. 
Therefore, the number of outputs is $\log_2 n$. Taking the 16-bit APC as an example, the output should be 4-bit to represent a number between 0 - 16. However,  it is worth noting that 
the weight of the least significant bit is $2^{1}$ rather than $2^{0}$ to represent 16. Therefore, the output of the APC is a bit-matrix with size of $\log_2 n \times m$. 

From Table \ref{tbl_apcinner}, we see that the APC-based inner product block only results in less than 1\% accuracy degradation when compared with the conventional accumulative parallel counter, but it can achieve about 40\% reduction of gate count \cite{kim2015approximate}. This observation demonstrates
the significant advantage of implementing 
efficient inner product block using APC-based method,
in terms of power, energy, and hardware resource.

\textbf{Two-Line Representation-Based Inner Product Block}.   
The two-line representation-based SC scheme~\cite{toral2000stochastic}
can be used to construct a non-scaled adder. Figure \ref{fig:sc_addition} (d) illustrates the structure of a two-line representation-based adder. Since $A_{i}$, $B_{i}$, and $C_{i}$ are bounded as the element of $\{-1, 0, 1\}$, a carry bit may be missed. Therefore, a three-state counter is used here to store the positive or negative carry bit. 

However, there are two limitations for the two-line representation-based inner product block in hardware DCNNs: {\em i)} An inner product block generally has more than two inputs, the overflow may often occur in the two-line representation-based inner product calculation due to its non-scaling characteristics. This leads to significant accuracy loss; and {\em ii)} the area overhead is too high compared with other inner product implementation methods.

\begin{table}[tbp]\footnotesize
\centering
\caption{Relative Errors of the APC-Based Compared with the Conventional Parallel Counter-Based Inner Product Blocks}
\label{tbl_apcinner}
\begin{tabular}{ccccc}
\hline
\multirow{2}{*}{Input size} & \multicolumn{4}{c}{\begin{tabular}[c]{@{}c@{}}Bit stream length\end{tabular}} \\ \cline{2-5} & 128    & 256    & 384    & 512    \\ \hline
16                              & 1.01\% & 0.87\% & 0.88\% & 0.84\% \\ 
32                              & 0.70\% & 0.61\% & 0.58\% & 0.57\% \\ 
64                              & 0.49\% & 0.44\% & 0.44\% & 0.42\% \\ \hline
\end{tabular}
\end{table}

\subsection{Pooling Block Designs}

Pooling (or down-sampling) operations are performed by pooling function blocks in DCNNs to significantly reduce 
{\em i)} inter-layer connections; and 
{\em ii)} the number of parameters and computations in the network, meanwhile maintaining the translation invariance of the extracted features \cite{cs231n}. Average pooling and max pooling are two widely used pooling strategies. Average pooling is straightforward to implement in SC domain, while max pooling, which exhibits higher performance in general, 
requires more hardware resources. In order to overcome this challenge, 
we propose a novel hardware-oriented 
max pooling design with high performance 
and amenable to SC-based implementation. 

\textbf{Average Pooling Block Design}. Figure \ref{fig:neuronconcept} (b) shows how the feature map is average pooled with $2 \times 2$ filters. Since average pooling calculates the mean value of entries in a small matrix, the inherent down-scaling property of the MUX can be utilized. Therefore, the average pooling can be performed by the structure shown in Figure \ref{fig:sc_addition} (b) with low hardware cost.


\textbf{Hardware-Oriented Max Pooling Block Design}. The max pooling operation 
has been recently shown to provide higher performance in practice 
compared with the average pooling operation \cite{cs231n}. 
However, in SC domain, we can find out the bit-stream with the maximum value among 
four candidates only after counting the total number of 1's through the whole bit-streams, 
which inevitably incurs long latency and considerable energy consumption. 

To mitigate the cost, we propose a novel SC-based hardware-oriented max pooling scheme. 
The insight is that once a set of bit-streams are sliced into segments, 
the globally largest bit-stream (among the four candidates) has the highest probability 
to be the locally largest one in each set of bit-stream segments. 
This is because all 1's are randomly distributed in the stochastic bit-streams. 

Consider the input bit-streams of the hardware-oriented max pooling block as a bit matrix. 
Suppose there are four bit-streams, and each has $m$ bits, thus the size of the bit matrix is $4 \times m$. 
Then the bit matrix is evenly sliced into small matrices whose size are $c \times m$ 
(i.e., each bit-stream is evenly sliced into segments whose length are $c$). 
Since the bit-streams are randomly generated, ideally, the largest row (segment) among the four rows 
in each small matrix is also the largest row of the global matrix. 
To determine the largest row in a small matrix, the number of 1s are counted in all rows 
in a small matrix in parallel.
The maximum counted result determines the next $c$-bit 
row that is sent to the output of the pooling block. 
In another word, the currently selected $c$-bit segment 
is determined by the counted results of the previous matrix. 
To reduce latency, the $c$-bit segment from the first small matrix is randomly chosen.
This strategy incurs zero extra latency but only causes 
a {\em negligible accuracy loss} when $c$ is properly selected.  

\begin{figure}[t]
	\centering
	\includegraphics[width=0.9\columnwidth]{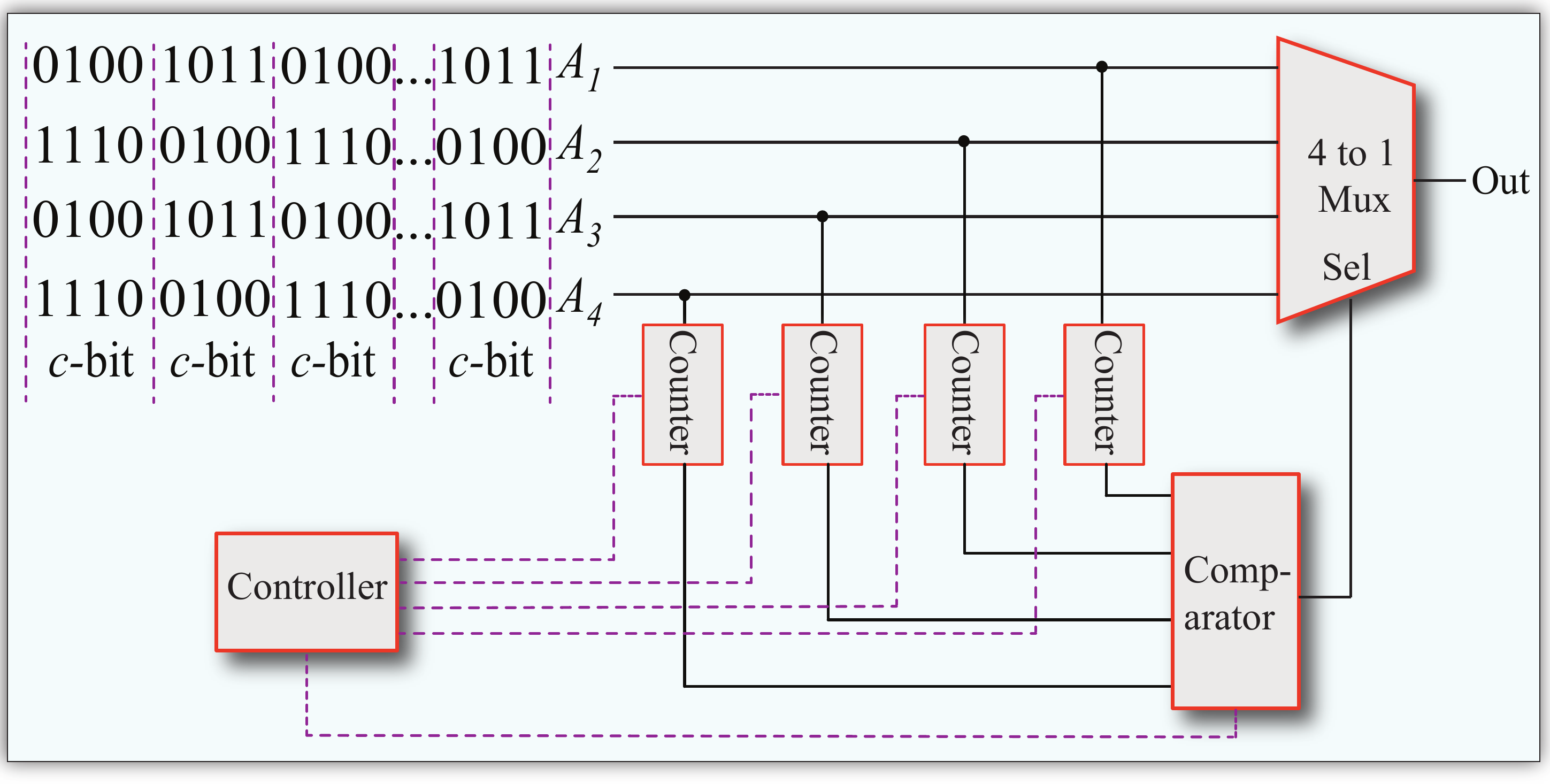}
	\caption{The Proposed Hardware-Oriented Max Pooling.}
	\label{fig:max_pooling}
\end{figure}

Figure \ref{fig:max_pooling} illustrates the structure of the hardware-oriented max pooling block, where the output from $max\_output$ approximately is equal to the largest bit-stream. 
The four input bit-streams sent to the multiplexer are also connected to four counters, and the outputs of the counters are connected to a comparator to determine the largest segment. 
The output of the comparator is used to control the selection of the four-to-one MUX. 
Suppose in the previous small bit matrix, the second row is the largest, 
then MUX will output the second row of the current small matrix as the current $c$-bit output.

Table \ref{tbl_maxpooling} shows the result deviations of the hardware-oriented max pooling design compared with the software-based max pooling implementation. The length of a bit-stream segment is 16. In general, the proposed pooling block can provide a sufficiently accurate result even with large input size.

\begin{table}[tbp]\footnotesize
\centering
\caption{Relative Result Deviation of Hardware-Oriented Max Pooling Block Compared with Software-Based Max Pooling}
\label{tbl_maxpooling}
\begin{tabular}{ccccc}
\hline
\multirow{2}{*}{Input size} & \multicolumn{4}{c}{Bit-stream length} \\\cline{2-5}
& 128   & 256   & 384   & 512   \\ \hline
4                               & 0.127 & 0.081 & 0.066 & 0.059 \\ 
9                               & 0.147 & 0.099 & 0.086 & 0.074 \\ 
16                              & 0.166 & 0.108 & 0.097 & 0.086 \\ \hline
\end{tabular}
\end{table}
 
\subsection{Activation Function Block Designs} \label{activation_unit}

\textbf{Stanh}. \cite{brown2001stochastic} proposed a $K$-state FSM-based design (i.e., Stanh) in the SC domain for implementing the tanh function and describes the relationship between Stanh and tanh as $Stanh(K, x) \cong tanh(\frac{K}{2}x)$. When the input stream $x$ is distributed in the range [-1, 1] (i.e. $\frac{K}{2}x$ is distributed in the range $[-\frac{K}{2}, \frac{K}{2}]$), this equation works well, and higher accuracy can be achieved with the increased state number $K$. 

\begin{figure}[t]
	\centering
	\includegraphics[width=0.75\columnwidth]{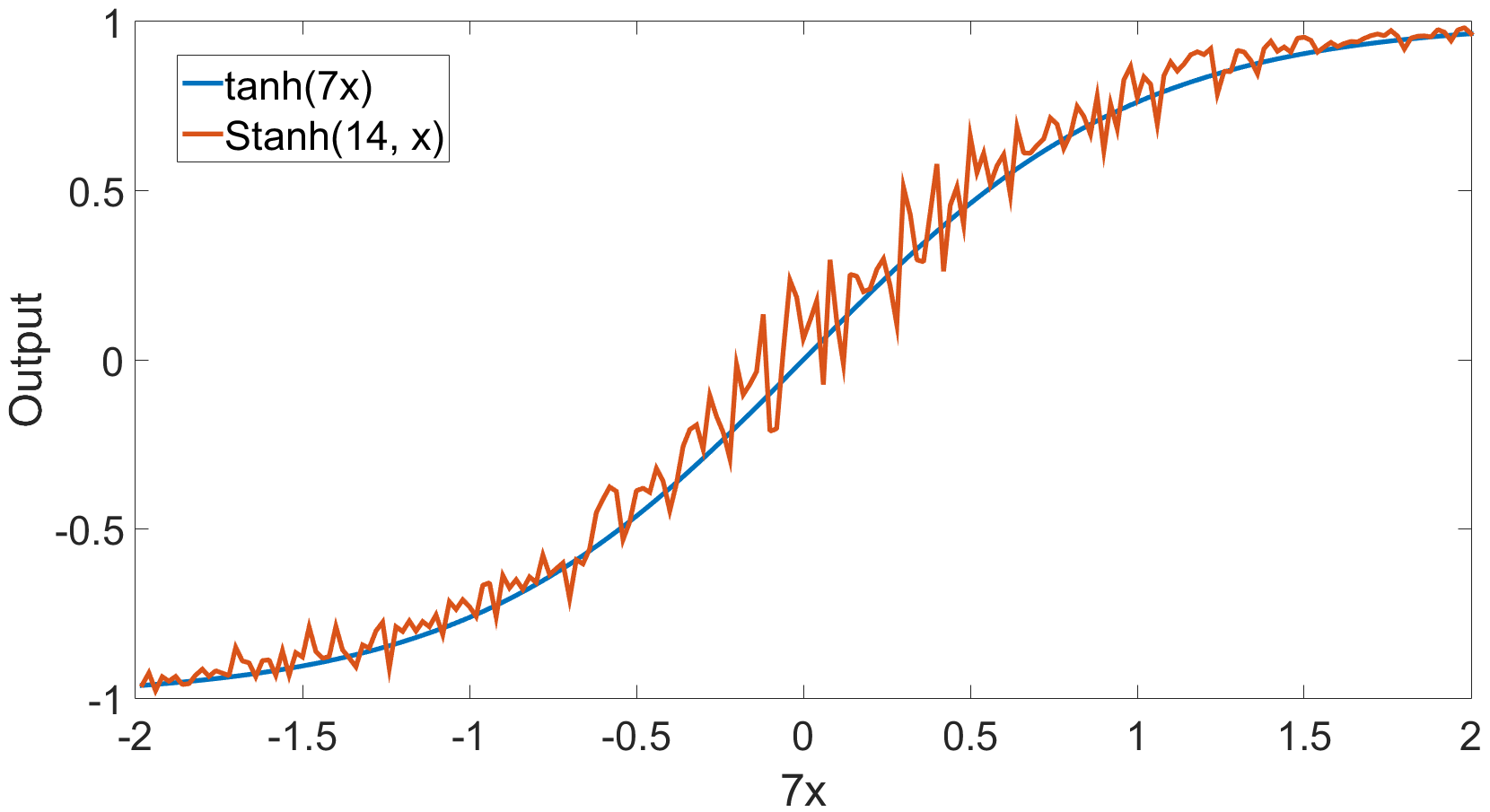}
    \vskip -0.2em
	\caption{Output comparison of Stanh vs tanh.}
	\label{fig:stanh_result}
\end{figure}

However, $Stanh$ cannot be applied directly in SC-DCNN for three reasons. 
First, as shown in Figure \ref{fig:stanh_result} and Table \ref{tbl_stanh} 
(with bit-stream length fixed at 8192), when the input variable of Stanh 
(i.e. $\frac{K}{2}x$) is distributed in the range [-1, 1], 
the inaccuracy is quite notable and is not suppressed with the increasing of $K$. 
Second, the equation works well when $x$ is precisely represented. 
However, when the bit-stream is not impractically long 
(less than $2^{16}$ according to our experiments), 
the equation should be adjusted with a consideration of bit-stream length. 
Third, in practice, we usually need to proactively down-scale 
the inputs since a bipolar stochastic number cannot reach beyond the range [-1, 1]. 
Moreover, the stochastic number may be sometimes passively down-scaled by certain components, 
such as a MUX-based adder or an average pooling block \cite{listructural,li2016dscnn}. 
Therefore, a \emph{scaling-back} process is imperative to obtain an accurate result. 
Based on the these reasons, the design of Stanh needs to be optimized 
together with other function blocks to achieve high accuracy for different bit-stream lengths 
and meanwhile provide a scaling-back function.
More details are discussed in Section \ref{feature extraction block}.

\begin{table}[tbp]\footnotesize
\centering
\caption{The Relationship Between State Number and Relative Inaccuracy of Stanh}
\label{tbl_stanh}
\resizebox{0.99\columnwidth}{!}{
\begin{tabular}{cccccccc}
\hline
State Number   & 8       & 10     & 12     & 14     & 16     & 18     & 20     \\ \hline
Relative Inaccuracy ($\%$) & 10.06 & 8.27 & 7.43 & 7.36 & 7.51 & 8.07 & 8.55 \\ \hline
\end{tabular}
}
\end{table}

\textbf{Btanh}. Btanh is specifically designed for the APC-based adder to perform a scaled hyperbolic tangent function. 
Instead of using FSM, a saturated up/down counter is used to convert the binary outputs of the APC-based adder 
back to a bit-stream. The implementation details and how to determine 
the number of states can be found in \cite{kim2016dynamic}.

\subsection{Design \& Optimization for Feature Extraction Blocks} \label{feature extraction block}

\begin{figure}[t]
	\centering
	\includegraphics[width=0.55\columnwidth]{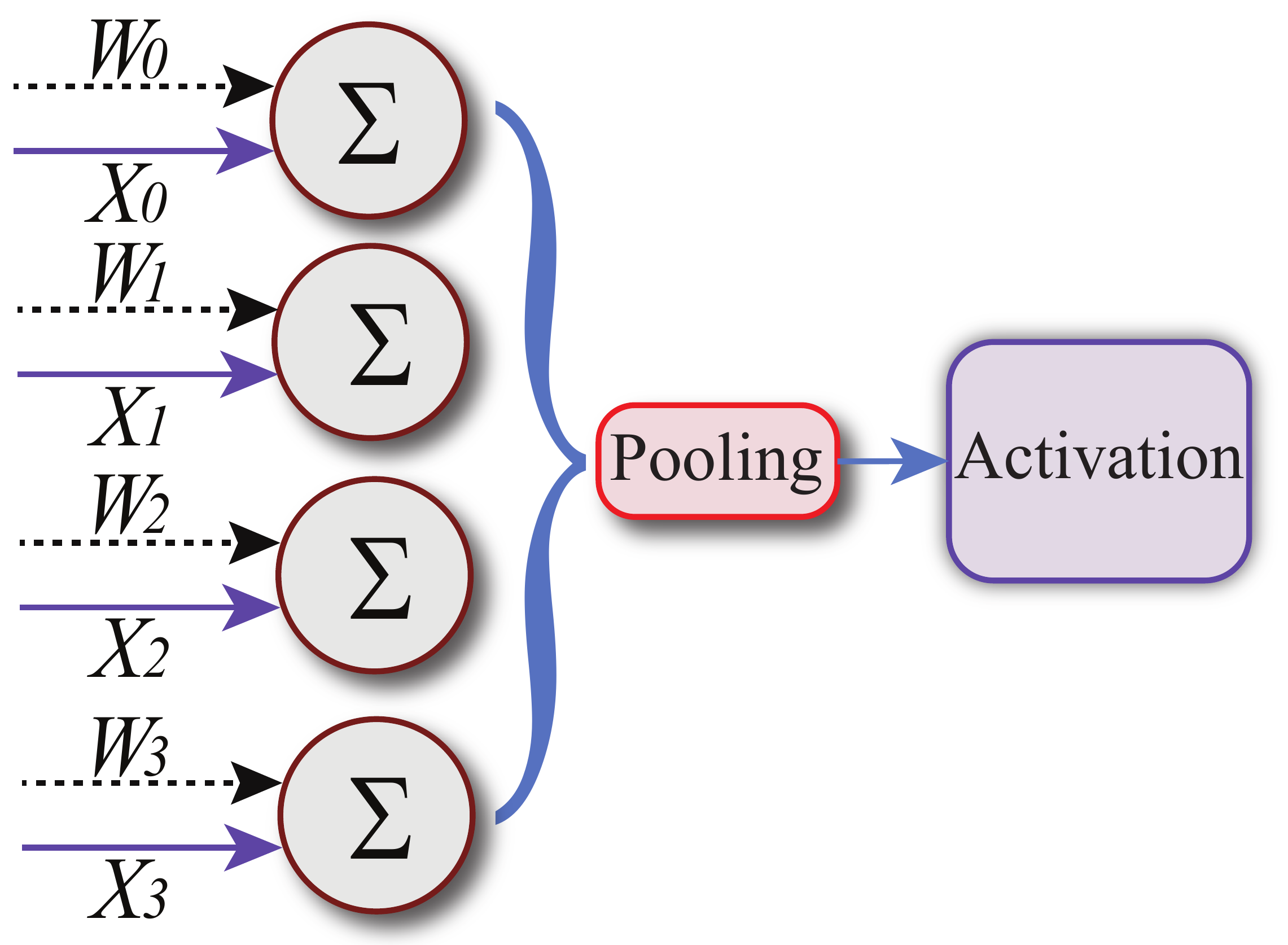}
	\caption{The structure of a feature extraction block.}
	\label{fig:featureblock}
\end{figure}

In this section, we propose an optimized feature extraction blocks.
Based on the previous analysis and results, we select several candidates for constructing 
feature extraction blocks shown in Figure \ref{fig:featureblock}, including:
the MUX-based and APC-based inner product/convolution blocks,
average pooling and hardware-oriented max pooling blocks, and 
Stanh and Btanh blocks.

In SC domain, the parameters such as input size, bit-stream length, and the inaccuracy introduced by the previous connected block can {\em collectively} affect the overall performance of the feature extraction block. 
Therefore, the isolated optimizations on each individual basic function block are insufficient 
to achieve the satisfactory performance for the entire feature extraction block. 
For example, the most important advantage of the APC-based inner product block 
is its high accuracy and thus the bit-stream length can be reduced.
On the other side, the most important advantage of MUX-based inner product block 
is the low hardware cost and the accuracy can be improved by increasing the bit-stream length. 
Accordingly, to achieve good performance, we cannot simply compose these basic function blocks,
instead, a series of joint optimizations are performed on each type of feature extraction block.
Specifically, we 
attempt to fully making use of the advantages of each of the building blocks.

In the following discussion, 
we use MUX/APC to denote the MUX-based or APC-based inner product/convolution blocks; 
use Avg/Max to denote the average or hardware-oriented max pooling blocks; 
use Stanh/Btanh to denote the corresponding activation function blocks. 
A feature extraction block configuration is represented by choosing various combinations
from the three components. 
For example, MUX-Avg-Stanh means that four MUX-based inner product blocks, 
one average pooling block, and one Stanh activation function block are cascade-connected
to construct an instance of feature extraction block.

\textbf{MUX-Avg-Stanh}. 
As discussed in Section \ref{activation_unit}, when Stanh is used, 
the number of states needs to be carefully selected with a comprehensive 
consideration of the scaling factor, bit-stream length, and accuracy requirement. 
Below is the empirical equation that is extracted from our comprehensive experiments to obtain the approximately optimal state number $K$ to achieve a high accuracy:
\small
\begin{equation}
\begin{aligned}
K = f(L,N) \approx 2 \times \log_2{N} + \frac{\log_2{L} \times N}{\alpha \times \log_2{N} } , \\
\end{aligned}
\end{equation}
\normalsize
where the nearest even number to the result calculated by the above equation is assigned to $K$, 
$N$ is the input size, $L$ is the bit-stream length, and empirical parameter $\alpha = 33.27$.

\begin{figure}[t]
	\centering
	\includegraphics[width=0.9\columnwidth]{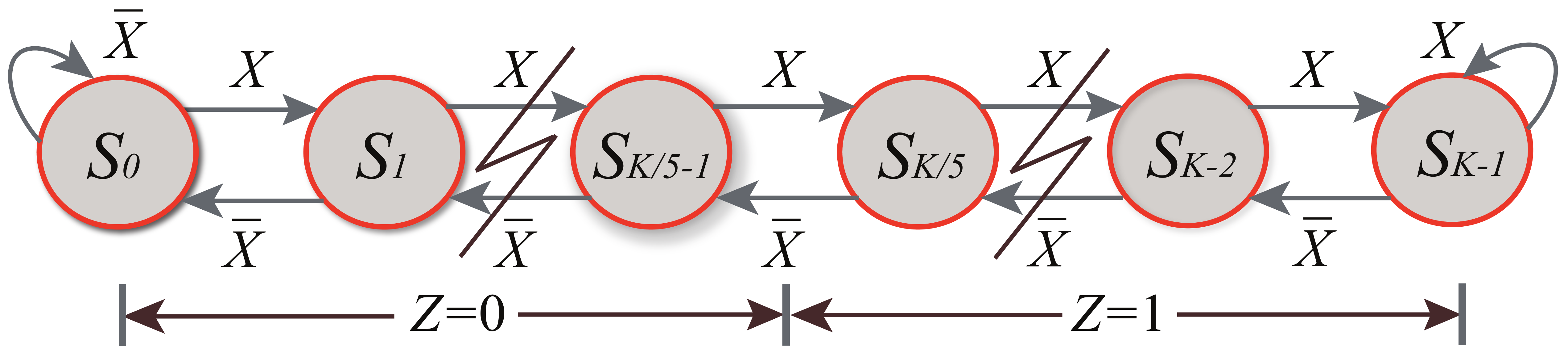}
    \vskip -0.4em
	\caption{Structure of optimized Stanh for MUX-Max-Stanh.}
	\label{fig:mux_max_fsm}
\end{figure}

\textbf{MUX-Max-Stanh}. The hardware-oriented max pooling block shown in Figure \ref{fig:max_pooling} 
in most cases generates an output that is slightly less than the maximum value. 
In this design of feature extraction block, the inner products are all scaled down by a factor of $n$ ($n$ is the input size), and the subsequent scaling back function of Stanh will enlarge the inaccuracy, especially when the positive/negative sign of the selected maximum inner product value is changed. For example, 
505/1000 is a positive number, and 1\% under-counting will lead the output of the hardware-oriented max pooling unit to be 495/1000, which is a negative number. Thereafter, the obtained output of Stanh may be -0.5, but the expected result should be 0.5. Therefore, the bit-stream has to be long enough to diminish the impact of under-counting, and the Stanh needs to be re-designed to fit the correct (expected) results. As shown in Figure \ref{fig:mux_max_fsm}, the re-designed FSM for Stanh will output 0 when the current state is at the left 1/5 of the diagram, otherwise it outputs a 1. 
The optimal state number $K$ is calculated through the following empirical equation derived from experiments: 
\small
\begin{equation}
K = f(L,N) \approx 2 \times (\log_2{N} + \log_2{L}) - \frac{\alpha}{\log_2{N}} -  \frac{\beta}{\log_5{L} } ,\\ 
\end{equation}
\normalsize
where the nearest even number to the result calculated by the above equation is assigned to $K$, $N$ is the input size, $L$ is the bit-stream length, $\alpha = 37$, and empirical parameter $\beta = 16.5$.

\textbf{APC-Avg-Btanh}. 
When the APC is used to construct the inner product block, conventional arithmetic calculation components, such as full adders and dividers, can be utilized to perform the averaging calculation, because the output of APC-based inner product block is a binary number. Since the design of Btanh initially aims at directly connecting to the output of APC, and an average pooling block is now inserted between APC and Btanh, the original formula proposed in \cite{kim2016dynamic} for calculating the optimal state number of Btanh needs to be re-formulated as:
\small
\begin{equation}
K = f(N) \approx \frac{N}{2} ,
\end{equation}
\normalsize
from our experiments. In this equation $N$ is the input size, and the nearest even number to $\frac{N}{2}$ is assigned to $K$.

\textbf{APC-Max-Btanh}. Although the output of APC-based inner product block is a binary number, the conventional binary comparator cannot be directly used to perform max pooling. This is because the output sequence of APC-based inner product block is still a stochastic bit-stream. If the maximum binary number is selected at each time, the pooling output is always greater than the actual maximum inner product result. Instead, the proposed hardware-oriented max pooling design should be used here, and the counters should be replaced by accumulators for accumulating the binary numbers. Thanks to the high accuracy provided by accumulators in selecting the maximum inner product result, the original Btanh design presented in \cite{kim2016dynamic} can be directly used without adjustment.   

\section{Weight Storage Scheme and Optimization}

As discussed in Section \ref{sctn3_neurondesign}, 
the main computing task of an inner product block is to calculate the inner products of $x_{i}$'s and $w_{i}$'s. 
$x_{i}$'s are input by users, and $w_{i}$'s are weights obtained by training using software and should be stored in the hardware-based DCNNs. 
\emph{Static random access memory (SRAM)} is the most suitable circuit structure for weight storage due to its high reliability, high speed, and small area. 
The specifically optimized SRAM placement schemes and weight storage methods are imperative for further reductions of area and power (energy) consumption. In this section, we present optimization techniques including efficient filter-aware SRAM sharing, weight storage method, and layer-wise weight storage optimizations.

\subsection{Efficient Filter-Aware SRAM Sharing Scheme}
 
\begin{figure}[t]
	\centering
	\includegraphics[width=0.75\columnwidth]{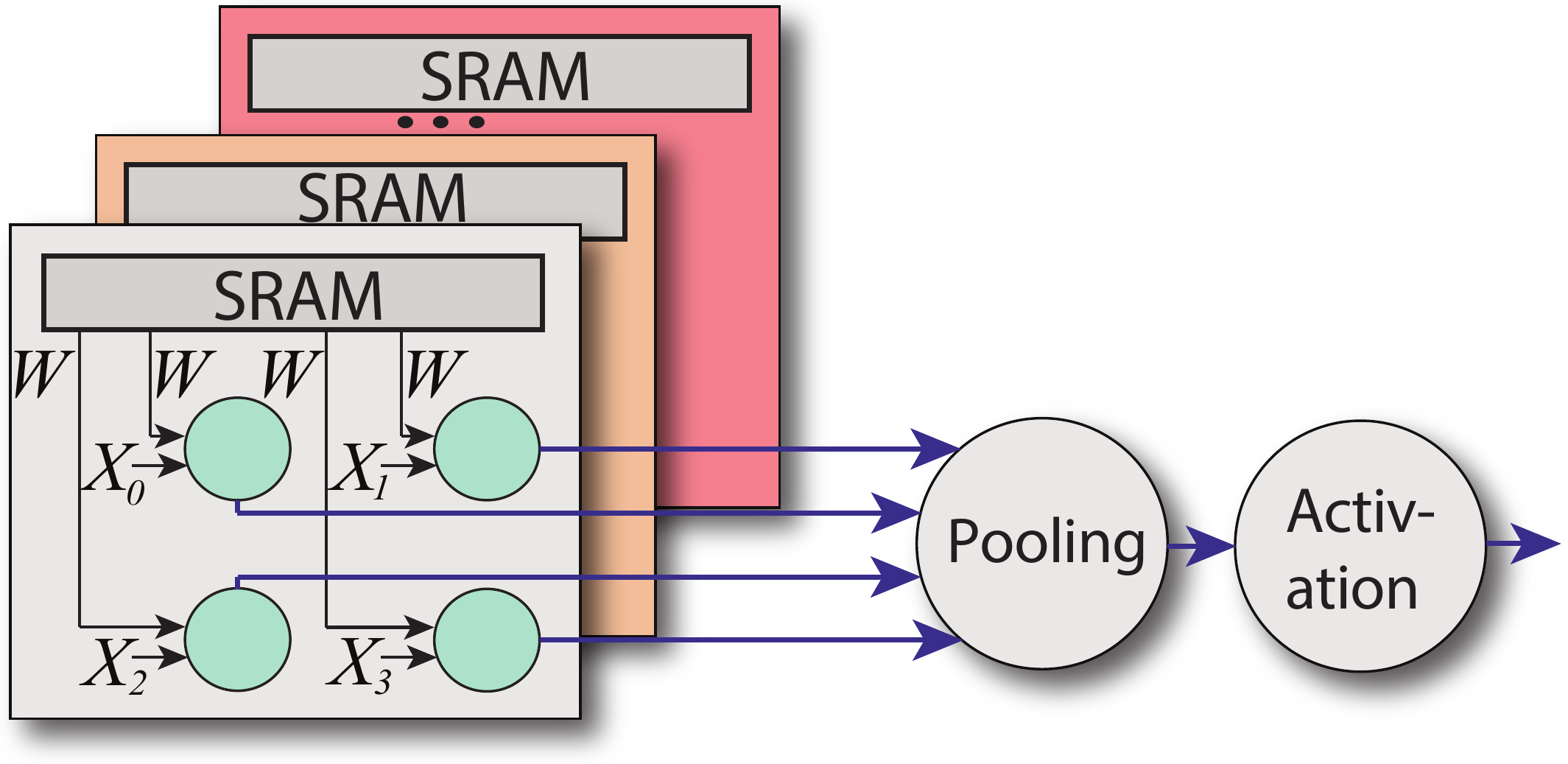}
	\caption{Filter-Aware SRAM Sharing Scheme.}
	\label{fig:sram_sharing}
\end{figure}

Since all receptive fields of a feature map share one filter (a matrix of weights), 
all weights functionally can be separated into filter-based blocks,
and each weight block is shared by all inner product/convolution blocks using the corresponding filter. 
Inspired by this fact, we propose an efficient filter-aware SRAM sharing scheme, 
with structure illustrated in Figure \ref{fig:sram_sharing}. 
The scheme divides the whole SRAM into small blocks to mimic filters. 
Besides, all inner product blocks can also be separated into feature map-based groups, 
where each group extracts a specific feature map. 
In this scheme, a local SRAM block is shared by all 
the inner product blocks of the corresponding group. 
The weights of the corresponding filter are stored in the local SRAM block of this group. 
This scheme significantly reduces the routing overhead and wire delay.

\subsection{Weight Storage Method} 
\vskip -0.3em

Besides the reduction on routing overhead, 
the size of SRAM blocks can also be reduced by trading off accuracy for less hardware resources. 
The trade-off is realized by eliminating certain least significant bits of a weight value to 
reduce the SRAM size. In the following, we present a weight storage method that significantly reduces 
the SRAM size with little network accuracy loss. 

\textbf{Baseline: High Precision Weight Storage}. 
In general, DCNNs are trained with single floating point precision. 
In hardware, the SRAM up to 64-bit is needed for storing one weight 
value in the fixed-point format to maintain its original high precision. 
This scheme provides high accuracy since there is almost no information loss of weights. 
However, it also incurs high hardware consumption in that the size of SRAM and its related 
read/write circuits increase with the precision increase of the stored weight values. 

\textbf{Low Precision Weight Storage Method}. 
According to our application-level experiments, 
many least significant bits that are far from the decimal point 
only have a very limited impact on the overall accuracy.
Therefore, the number of bits for weight representation in the SRAM block can be significantly reduced. 
We propose a mapping equation that converts a weight in the real number format to the binary number stored in SRAM to eliminate the proper numbers of least significant bits. 
Suppose the weight value is $x$, and the number of bits to store a weight value
in SRAM is $w$ (which is defined as the \emph{precision} of the represented
weight value), then the binary number to be stored for representing $x$ is:
$y=\frac{Int(\frac{x+1}{2} \times 2^w)}{2^w}$, where $Int()$ means only keeping
the integer part. Figure \ref{fig:layer_wise_precision} illustrates the network 
error rates when the reductions of weights' precision are performed at a single layer or all layers. 
The precision loss of weights at Layer0 (consisting of a convolutional layer and pooling layer) 
has the least impact, while the precision loss of weights at Layer2 (a fully connected layer) 
has the most significant impact. The reason is that Layer2 is the fully connected layer that has the largest number of weights. On the other hand, when $w$ is set equal to or greater than seven, 
the network error rates are low enough and almost do not decrease
 with the further precision increase. 
 Therefore, the proposed weight storage method can significantly 
 reduce the size of SRAMs and their read/write circuits by reducing precision. 
 The area savings estimated using CACTI 5.3 \cite{cacti} is 10.3$\times$.

\subsection{Layer-wise Weight Storage Optimization}
\begin{figure}[t]
	\centering
	\includegraphics[width=0.8\columnwidth]{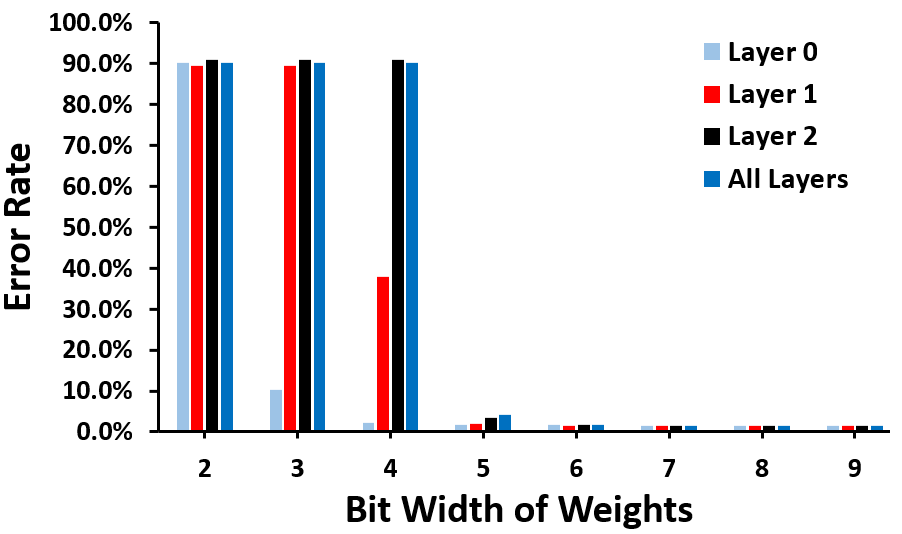}
    \vskip -0.2em
	\caption{The impact of precision of weights at different layers on the overall SC-DCNN network accuracy.}
	\label{fig:layer_wise_precision}
\end{figure}

As shown in Figure \ref{fig:layer_wise_precision}, the precision of weights at different layers have different impacts on the overall accuracy of the network. \cite{judd2015reduced} presented a method that set different weight precisions at different layers to save weight storage. In SC-DCNN, we adopt the same strategy to improve the hardware performance. 
Specifically, this method is effective to obtain savings in SRAM area and power (energy) consumption because Layer2 has the most number of weights compared with the previous layers. 
For instance, when we set weights as 7-7-6 at the three layers of LeNet5, 
the network error rate is 1.65\%, which has only 0.12\% accuracy degradation 
compared with the error rate obtained using software-only implementation. 
However, 12$\times$ improvements on area and 11.9$\times$ improvements on power consumption are achieved for the weight representations (from CACTI 5.3 estimations), compared with the baseline without any reduction in weight representation bits.
 

\begin{figure*}
	\centering
	\includegraphics[width=0.95\textwidth]{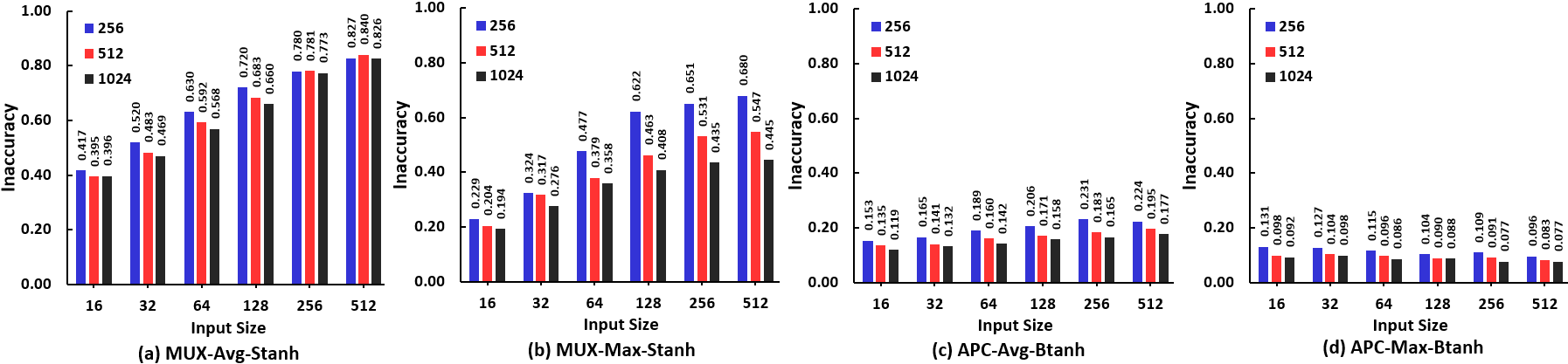}
   \vskip -0.2em
	\caption{Input size versus absolute inaccuracy for (a) MUX-Avg-Stanh, (b) MUX-Max-Stanh, (c) APC-Avg-Btanh, and (d) APC-Max-Btanh with different bit stream lengths.}
	\label{fig:exp_accuracy}
\end{figure*}

\begin{figure*}
	\centering
	\includegraphics[width=0.95\textwidth]{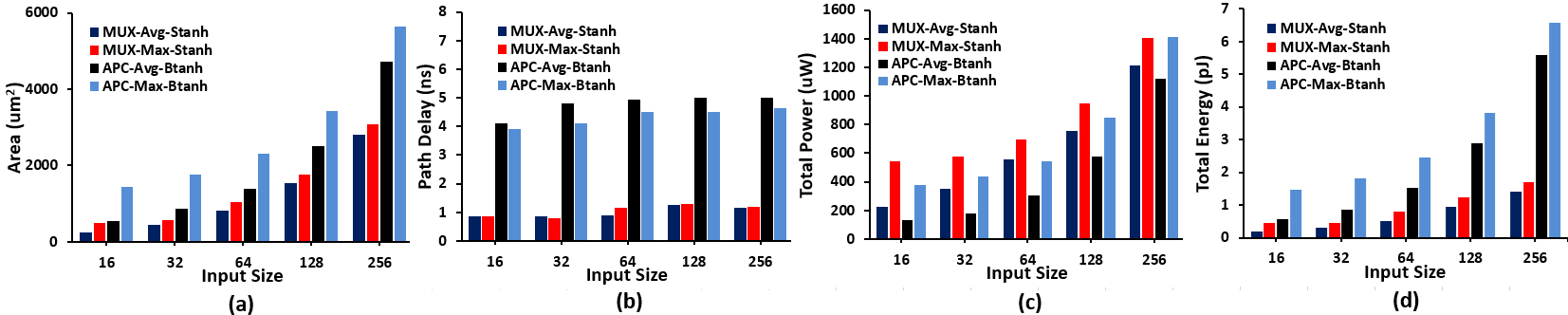}
   \vskip -0.2em
	\caption{Input size versus (a) area, (b) path delay, (c) total power, and (d) total energy for four different designs of feature extraction blocks.}
	\label{fig:exp_hardware}
\end{figure*}


\section{Overall SC-DCNN Optimizations and Results} \label{experiments}

In this section, we present optimizations of feature extraction blocks 
along with comparison results with respect to accuracy, area/hardware footprint, power (energy) consumption, etc. 
Based on the results, we perform thorough optimizations on the overall SC-DCNN 
to construct LeNet5 structure, which is one of the most well-known large-scale deep DCNN structure.
The goal is to minimize area and power (energy) consumption while maintaining a
high network accuracy. 
We present comprehensive comparison results among {\em i)} SC-DCNN designs with
different target network accuracy,
and {\em ii)} existing hardware platforms. 
The hardware performance of the various SC-DCNN implementations 
regarding area, path delay, power and energy consumption are obtained by: 
{\em i)} synthesizing with the 45nm Nangate Open Cell Library \cite{nangate45} using Synopsys Design Compiler; and
{\em ii)} estimating using CACTI 5.3 \cite{cacti} for the SRAM blocks. 
The key peripheral circuitries in the SC domain (e.g. the random number generators) are developed using the design in \cite{kim2016energy} and synthesized using Synopsys Design Compiler.

\vskip -0.25em
\subsection{Optimization Results on Feature Extraction Blocks}
\vskip -0.25em
We present optimization results of feature extraction blocks with 
different structures, input sizes, and bit-stream lengths on accuracy, area/hardware footprint, power (energy) consumption, etc. Figure \ref{fig:exp_accuracy} illustrates the accuracy results of four types of feature extraction blocks: MUX-Avg-Stanh, MUX-Max-Stanh, APC-Avg-Btanh, and APC-Max-Btanh. The horizontal axis represents the input size that increases logarithmically from 16 ($2^4$) to 256 ($2^8$). The vertical axis represents the hardware inaccuracy of feature extraction blocks. Three bit-stream lengths are tested and their impacts are shown in the figure. 
Figure \ref{fig:exp_hardware} illustrates the comparisons among four feature extraction blocks with respect to area, path delay, power, and energy consumption. 
The horizontal axis represents the input size that increases logarithmically from 16 ($2^4$) to 256 ($2^8$). The bit-stream length is fixed at 1024.

\textbf{MUX-Avg-Stanh}. 
From Figure \ref{fig:exp_accuracy} (a), we see that it has the worst accuracy performance among the four structures. It is because the MUX-based adder,
as mentioned in Section \ref{sctn3_neurondesign},
is a down-scaling adder and incurs inaccuracy due to information loss. 
Moreover, average pooling is performed with MUXes, thus the inner products are further down-scaled and more inaccuracy is incurred. 
As a result, this structure of feature extraction block is only appropriate for dealing with receptive fields with a small size. 
On the other hand, it is the most area and energy efficient design with the smallest path delay. 
Hence, it is appropriate for scenarios with tight limitations on area and delay.

\textbf{MUX-Max-Stanh}.
Figure \ref{fig:exp_accuracy} (b) shows that it has a better accuracy compared with MUX-Avg-Stanh. 
The reason is that the mean of four numbers is generally closer to zero than the maximum value of the four numbers. As discussed in Section \ref{sctn3_neurondesign}, minor inaccuracy on the stochastic numbers near zero can cause significant inaccuracy on the outputs of feature extraction blocks. Thus the structures with hardware-oriented pooling are more resilient than the structures with average pooling. In addition, the accuracy can be significantly improved by increasing the bit-stream length, thus this structure can be applied for dealing with the receptive fields with both small and large sizes. With respect to area, path delay, and energy, its performance is a close second to the MUX-Avg-Stanh. 
Despite its relatively high power consumption, 
the power can be remarkably reduced by trading off the path delay.

\textbf{APC-Avg-Btanh}.
Figure \ref{fig:exp_accuracy} (c) and \ref{fig:exp_accuracy} (d) illustrate the hardware inaccuracy
of APC-based feature extraction blocks. The results imply that they provide significantly 
higher accuracy than the MUX-based feature extraction blocks.
It is because the APC-based inner product blocks maintain 
most information of inner products and thus generate results with high accuracy.
It is exactly the drawback of the MUX-based inner product blocks. 
On the other hand, APC-based feature extraction blocks consume more hardware resources 
and result in much longer path delays and higher energy consumption. 
The long path delay is partially the reason why the power consumption is lower than MUX-based designs. Therefore, the APC-Avg-Btanh is appropriate for DCNN implementations that have a tight specification on the accuracy performance and have a relative loose hardware resource constraint.

\textbf{APC-Max-Btanh}.
Figure \ref{fig:exp_accuracy} (d) indicates that this feature extraction block design has the best accuracy due to several reasons. First, it is an APC-based design. Second, the average pooling in the APC-Avg-Btanh causes more information loss than the proposed hardware-oriented max pooling. To be more specific, the fractional part of the number after average pooling is dropped: the mean of (2, 3, 4, 5) is 3.5, but it will be represented as 3 in binary format, thus some information is lost during the average pooling. Generally, the increase of input size will incur significant inaccuracy except for APC-Max-Btanh. The reason that APC-Max-Btanh performs better with more inputs is: more inputs will make the four inner products sent to the pooling function block more distinct from one another, 
in another word, more inputs result in higher accuracy in selecting the maximum value. The drawbacks of APC-Max-Btanh are also different. 
It has the highest area and energy consumption, and its path delay is just second (but very close) to APC-Avg-Btanh. 
Also, its power consumption is second (but close) to the MUX-Max-Stanh. 
Accordingly, this design is appropriate for the applications that have a very tight requirement on the accuracy. 

\vskip -0.25em
\subsection{Layer-wise Feature Extraction Block Configurations}
\begin{figure}[t]
	\centering
	\includegraphics[width=0.9\columnwidth]{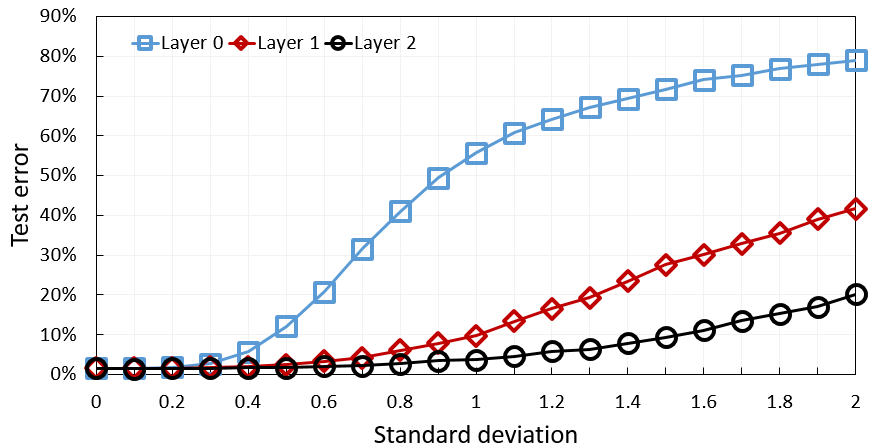}
	\vskip -0.2em
	\caption{The impact of inaccuracies at each layer on the overall SC-DCNN network accuracy.}
	\label{fig:layer_accuracy}
\end{figure}

\vskip -0.5em
Inspired by the experiment results of the layer-wise weight storage scheme, we also 
investigate the sensitivities of different layers to the inaccuracy.
Figure \ref{fig:layer_accuracy} illustrates that 
different layers have different error sensitivities. 
Combining this observation with the observations drawn from Figure
\ref{fig:exp_accuracy} and Figure \ref{fig:exp_hardware}, we propose a
layer-wise configuration strategy that uses different types of feature
extraction blocks in different layers to minimize area and power (energy)
consumption while maintaining a high network accuracy.

\begin{table*}\tiny
\centering
\caption{Comparison among Various SC-DCNN Designs Implementing LeNet 5}
\vskip -0.1em
\label{tbl_final1}
\resizebox{1\textwidth}{!}{
\begin{tabular}{cccccccccccc}
\hline
\multirow{2}{*}{ No. }&  \multirow{2}{*}{ Pooling } & Bit & \multicolumn{3}{c}{Configuration} & &\multicolumn{5}{c}{Performance} \\\cline{4-6}\cline{8-12}
&  & Stream & Layer 0 & Layer 1 & Layer 2 & & Inaccuracy (\%) & Area ($mm^2$) & Power ($W$) & Delay (ns)& Energy ($\mu J$)\\\hline
1 & \multirow{6}{*}{Max} & \multirow{2}{*}{1024} &MUX	&MUX	&APC	& &2.64	&19.1	&1.74
	&5120	&8.9\\
2 & & & MUX	&APC	&APC	& &2.23	&22.9	&2.13	&5120	&10.9\\
3 & & \multirow{2}{*}{512} & APC	&MUX	&APC	& &1.91	&32.7	&3.14	&2560	&8.0\\
4 & & & APC	&APC	&APC	& &1.68	&36.4	&3.53	&2560	&9.0\\
5 & & \multirow{2}{*}{256}& APC	&MUX	&APC	& &2.13	&32.7	&3.14	&1280	&4.0\\
6 & & & APC	&APC	&APC	& &1.74	&36.4	&3.53	&1280	&4.5\\\cline{2-12}
7 & \multirow{6}{*}{Average} & \multirow{2}{*}{1024} &MUX	&APC	&APC & 	&3.06	&17.0	&1.53	&5120	&7.8\\
8 & & & APC	&APC	&APC	& &2.58	&22.1	&2.14	&5120	&11.0\\
9 & & \multirow{2}{*}{512}& MUX	&APC	&APC	& &3.16	&17.0	&1.53	&2560	&3.9\\
10 & & & APC	&APC	&APC	& &2.65	&22.1	&2.14	&2560	&5.5\\
11 & & \multirow{2}{*}{256}& MUX	&APC	&APC	& &3.36	&17.0	&1.53	&1280	&2.0\\
12 & & & APC	&APC	&APC	& &2.76	&22.1	&2.14	&1280	&2.7\\\hline
\end{tabular}
}
\end{table*}

\begin{table*}\footnotesize
\centering
\caption{List of Existing Hardware Platforms}
\label{tbl_final2}
\vskip -0.1em
\resizebox{1\textwidth}{!}{
\begin{tabular}{ccccccccccc}
\hline
\multirow{2}{*}{ Platform} & \multirow{2}{*}{Dataset} & Network  &\multirow{2}{*}{ Year} & Platform  & Area & Power & Accuracy & Throughput & Area Efficiency & Energy Efficiency\\
& & Type &  & Type & ($mm^2$) & (W) & (\%) & (Images/s) & (Images/s/$mm^2$) & (Images/J)\\\hline
\textbf{SC-DCNN (No.6)} & \multirow{7}{*}{MNIST} & \multirow{4}{*}{CNN}	& 2016 &	ASIC &36.4	&3.53	&98.26	&781250	&21439	&221287 \\
\textbf{SC-DCNN (No.11)} &	& &2016 &	ASIC		&17.0	&1.53	& 96.64	&781250	&45946	&510734 \\
2$\times$Intel Xeon W5580 &  & 	& 2009&	CPU	&	263&	156& 98.46	& 656& 2.5& 4.2\\
Nvidia Tesla C2075&	& & 2011&	GPU&		520&	202.5&	98.46	& 2333& 4.5& 3.2\\ \cline{3-3}
Minitaur \cite{Minitaur}& & ANN$^1$  & 2014&	FPGA&	N/A	& $\leq$1.5& 92.00		& 4880 & N/A& $\geq$3253\\\cline{3-3}
SpiNNaker \cite{SpiNNaker}& & DBN$^2$ &	2015&	ARM&	N/A	&	0.3&	95.00& 50& N/A & 166.7\\	\cline{3-3}
TrueNorth \cite{TrueNorth,TrueNorth1} &	& SNN$^3$ & 2015 &	ASIC 		&430	&0.18	& 99.42	& 1000	&2.3	&9259 \\\cline{2-3}
DaDianNao \cite{chen2014dadiannao} & \multirow{2}{*}{ImageNet} & CNN	& 2014 &	ASIC		&67.7	&15.97	&N/A	&147938	&2185	&9263 \\
EIE-64PE \cite{EIE} &  & CNN layer & 	2016 &	ASIC	&40.8	&0.59	&N/A	&81967	&2009	&138927 \\\hline
\end{tabular}
}
\end{table*}

\vskip -0.35em
\subsection{Overall Optimizations and Results on SC-DCNNs}
\vskip -0.5em

Using the design strategies presented so far, 
we perform holistic optimizations on the overall SC-DCNN to construct the LeNet 5 DCNN structure. The (max pooling-based or average pooling-based) LeNet 5 is a widely-used DCNN structure \cite{lecun2015lenet} with a configuration of 784-11520-2880-3200-800-500-10. The SC-DCNNs are evaluated with the MNIST handwritten digit image dataset \cite{deng2012mnist}, which consists of 60,000 training data and 10,000 testing data.

The baseline error rates of the max pooling-based and average pooling-based LeNet5 DCNNs using software implementations are 1.53\% and 2.24\%, respectively. In the optimization procedure, we set 
1.5\% as the threshold on the error rate difference compared with the error rates of software
implementation.
In another word, the network accuracy degradation of the SC-DCNNs cannot exceed 1.5\%. 
We set the maximum bit-stream length as 1024 to avoid excessively long delays. 
In the optimization procedure, for the configurations that achieve the target network accuracy, 
the bit-stream length is reduced by half in order to reduce energy consumption. Configurations are removed if they fail to meet the network accuracy goal. The process is iterated until no configuration is left. 

Table 6 displays some selected typical configurations and their comparison results (including the consumption of SRAMs and random number generators). Configurations No.1-6 are max pooling-based SC-DCNNs, and No.7-12 are average pooling-based SC-DCNNs. It can be observed that the configurations involving more MUX-based feature extraction blocks achieve lower hardware cost.
Those involving more APC-based feature extraction blocks achieve higher accuracy. 
For the max pooling-based configurations, No.1 is the most area efficient as well as power efficient configuration, and No.5 is the most energy efficient configuration. With regard to the average pooling-based configurations, No.7, 9, 11 are the most area efficient and power efficient configurations, and No.11 is the most energy efficient configuration.
\let\thefootnote\relax\footnote{\vskip -3em \noindent\rule{\columnwidth}{0.4pt}\\\noindent $^1$ANN: Artificial Neural Network; $^2$DBN: Deep Belief Network; $^3$SNN: Spiking Neural Network}

Table 7 shows the results of two configurations of our proposed SC-DCNNs together 
with other software and hardware platforms. 
It includes software implementations using Intel Xeon Dual-Core W5580 and Nvidia Tesla C2075 GPU 
and five hardware platforms: 
Minitaur \cite{Minitaur},
SpiNNaker \cite{SpiNNaker},
TrueNorth \cite{TrueNorth,TrueNorth1},
DaDianNao \cite{chen2014dadiannao}, and
EIE-64PE \cite{EIE}.
EIE's performance was evaluated on a fully connected layer of AlexNet \cite{krizhevsky2012imagenet}. The state-of-the-art platform DaDianNao proposed an ASIC ``node'' that could be connected in parallel to implement a large-scale DCNN.
Other hardware platforms implement different types of hardware neural networks such as spiking neural network or deep-belief network.  

For SC-DCNN, the configuration No.6 and No.11 are selected to compare with software implementation on CPU server or GPU.
No.6 is selected because it is the most accurate max pooling-based configuration.
No.11 is selected because it is the most energy efficient average pooling-based configuration. According to Table 7, the proposed SC-DCNNs are much more area efficient:
the area of Nvidia Tesla C2075 is up to 30.6$\times$ of the area of SC-DCNN (No.11).
Moreover, our proposed SC-DCNNs also have outstanding performance in terms of throughput, area efficiency, and energy efficiency. Compared with Nvidia Tesla C2075,
the SC-DCNN (No.11) achieves 15625$\times$ throughput improvements and 159604$\times$ energy efficiency improvements.

Although LeNet5 is relatively small, it is still a representative DCNN. 
Most of the computation blocks of LeNet5 can be applied to other networks (e.g. AlexNet). 
Based on our experiments, inside a network, the inaccuracy introduced by the SC-based components can significantly compensate each other. Therefore, we expect that SC-induced 
inaccuracy will be further reduced with larger networks. 
We also anticipate higher area/energy efficiency in larger DCNNs. 
Many of the basic computations are the same for different types of networks, thus the significant area/energy efficiency improvement in each component will result in improvement of the whole network (compared with binary designs) for different network sizes/types. In addition, when the network is larger, there are potentially more optimization space for further improving the area/energy efficiency. Therefore, we claim that the proposed SC-DCNNs have good scalability.
The investigations on larger networks will be conducted in our future work.   

\vskip -0.5em
\section{Conclusion}
\vskip -0.5em

In this paper, we propose {\em SC-DCNN}, 
the first comprehensive design and optimization framework of SC-based DCNNs.
SC-DCNN fully utilizes the advantages of SC and achieves remarkably low hardware footprint, 
low power and energy consumption, while maintaining high network accuracy. 
We fully explore the design space of different components to 
achieve high power (energy) efficiency and low hardware footprint. 
First, we investigated various function blocks including inner product calculations, pooling operations, and activation functions. 
Then we propose four designs of feature extraction blocks, which are in charge of extracting features from input feature maps, by connecting different basic function blocks with joint optimization.
Moreover, three weight storage optimization schemes are investigated for reducing the area and power (energy) consumption of SRAM. Experimental results demonstrate that our proposed SC-DCNN achieves low hardware footprint and low energy consumption. It achieves the throughput of 781250 images/s, area efficiency of 45946 images/s/$mm^2$, and energy efficiency of 510734 images/J. 

\section{Acknowledgement}
We thank anonymous reviewers for their valuable feedback.
This work is funded in part by the seedling fund of DARPA SAGA program under FA8750-17-2-0021. Besides, this work is also supported by Natural Science Foundation of China (61133004, 61502019) and
Spanish Gov. \& European ERDF under TIN2010-21291-C02-01 and Consolider CSD2007-00050. 

\bibliographystyle{abbrvnat}
\bibliography{references}

\end{document}